\documentclass[journal]{IEEEtran}

\usepackage{url}
\usepackage{color}
\usepackage{amsmath}
\usepackage{array,graphicx,subfigure}
\usepackage{multirow}

\usepackage{colortbl}

\usepackage{amsfonts}
\usepackage{hyperref}
\hypersetup{
    colorlinks=true,
    linkcolor=black,
    filecolor=black,      
    urlcolor=black,
    citecolor=black,
}

\usepackage{algorithm}
\usepackage{algpseudocode}

\newcommand{\tabincell}[2]{\begin{tabular}{@{}#1@{}}#2\end{tabular}}

\newcommand{\cut}[1]{}

\newcommand{\etal}{\textit{et al}.}
\newcommand{\ie}{\textit{i}.\textit{e}.}
\newcommand{\eg}{\textit{e}.\textit{g}.}
\newcommand{\etc}{\textit{etc}}

\newcommand{\red}[1]{\textcolor{red}{#1}}

\newcommand{\blue}[1]{\textcolor{blue}{#1}}
\newcommand{\black}[1]{\textcolor{black}{#1}}

\newcommand{\magenta}[1]{\textcolor{magenta}{#1}}
\ifCLASSINFOpdf
  
\else
 
\fi

\begin{document}

\title{Multi-Graph Transformer \\ for Free-Hand Sketch Recognition}
%\title{Multi-Graph Transformer: A Novel Graph Neural Network for Free-Hand Sketch}

\author{Peng~Xu$^{*}$,~~Chaitanya~K.~Joshi,~~and~~Xavier~Bresson
\thanks{\href{http://www.pengxu.net}{Peng Xu}, \href{https://scholar.google.com/citations?hl=en&user=cwxVFVgAAAAJ}{Chaitanya~K.~Joshi}, and \href{https://scholar.google.com/citations?user=9pSK04MAAAAJ&hl=zh-CN}{Xavier~Bresson} are with School of Computer Science and Engineering, Nanyang Technological University, Singapore.}% <-this % stops a space
\thanks{$^{*}$~Corresponding to Peng Xu, email: \url{peng.xu@ntu.edu.sg}.}
\thanks{Xavier Bresson is supported in part by NRF Fellowship NRFF2017-10.}
%\thanks{We would thank Prof. \href{https://scholar.google.com/citations?user=8kzzUboAAAAJ&hl=zh-CN}{Liang Wang} for his valuable suggestion that evaluabtes our multi-graph transformer idea on the Relation Extraction task.}
}

%\markboth{Journal of \LaTeX\ Class Files,~Vol.~a, No.~b, August~2015}%

\maketitle

\begin{abstract}
Learning meaningful representations of free-hand sketches remains a challenging task given the  signal sparsity and the high-level abstraction of sketches. Existing techniques have focused on exploiting either the static nature of sketches with Convolutional Neural Networks (CNNs) or the temporal sequential property with Recurrent Neural Networks (RNNs). 
In this work, we propose a new representation of sketches as multiple sparsely connected graphs.
We design a novel Graph Neural Network (GNN), the Multi-Graph Transformer~(MGT), for learning representations of sketches from multiple graphs, which simultaneously capture global and local geometric stroke structures, as well as temporal information.
\black{
We report extensive numerical experiments on a sketch recognition task to demonstrate the performance of the proposed approach. 
Particularly, MGT applied on 414k sketches from Google QuickDraw:
(i) achieves small recognition gap to the CNN-based performance upper bound ($72.80\%$ vs. $74.22\%$) and infers faster than the CNN competitors,
%yet outperforms most traditional CNNs, 
and
(ii) outperforms all RNN-based models by a significant margin.
}
To the best of our knowledge, this is the first work~\footnote{{The preliminary version of this work can be found in arXiv~\url{https://arxiv.org/abs/1912.11258v1} and \url{https://arxiv.org/abs/1912.11258v2}}} proposing to represent sketches as graphs and apply GNNs for sketch recognition.
{Code and trained models are available at
\url{https://github.com/PengBoXiangShang/multigraph_transformer}.}
\end{abstract}


%\IEEEPARstart{D}{eep}

\IEEEpeerreviewmaketitle

\section{Introduction}
\IEEEPARstart{F}{ree-hand} sketches are drawings made without the use of any instruments. 
Sketches are different from traditional images: they are formed of temporal sequences of strokes~\cite{ha2017sketchrnn, xu2018sketchmate}, while  images are static collections of pixels with dense color and texture patterns. Sketches capture high-level abstraction of visual objects with very sparse information compared to regular images, which makes the modelling of sketches unique and challenging.

The modern prevalence of touchscreen devices has led to a flourishing of sketch-related applications in recent years, including sketch recognition~\cite{liu2019sketchgan,sarvadevabhatla2016enabling}, sketch scene understanding~\cite{ye2016human}, sketch hashing~\cite{xu2018sketchmate}, sketch-based image retrieval~\cite{sangkloy2016sketchy,liu2017deep,shen2018zero,Collomosse_2019_CVPR,Dutta_2019_CVPR,Dey_2019_CVPR},   sketch-related generation~\cite{ha2017sketchrnn,Chen_2018_CVPR,lu2018image,liu2019sketchgan}, {sketch self-supervised learning~\cite{9119480,lin2020sketch}, \etc.}

\begin{figure}[!t]
	\centering
	\subfigure[original sketch]{
		\label{fig:cat_3_a}
		\includegraphics[width=0.31\columnwidth]{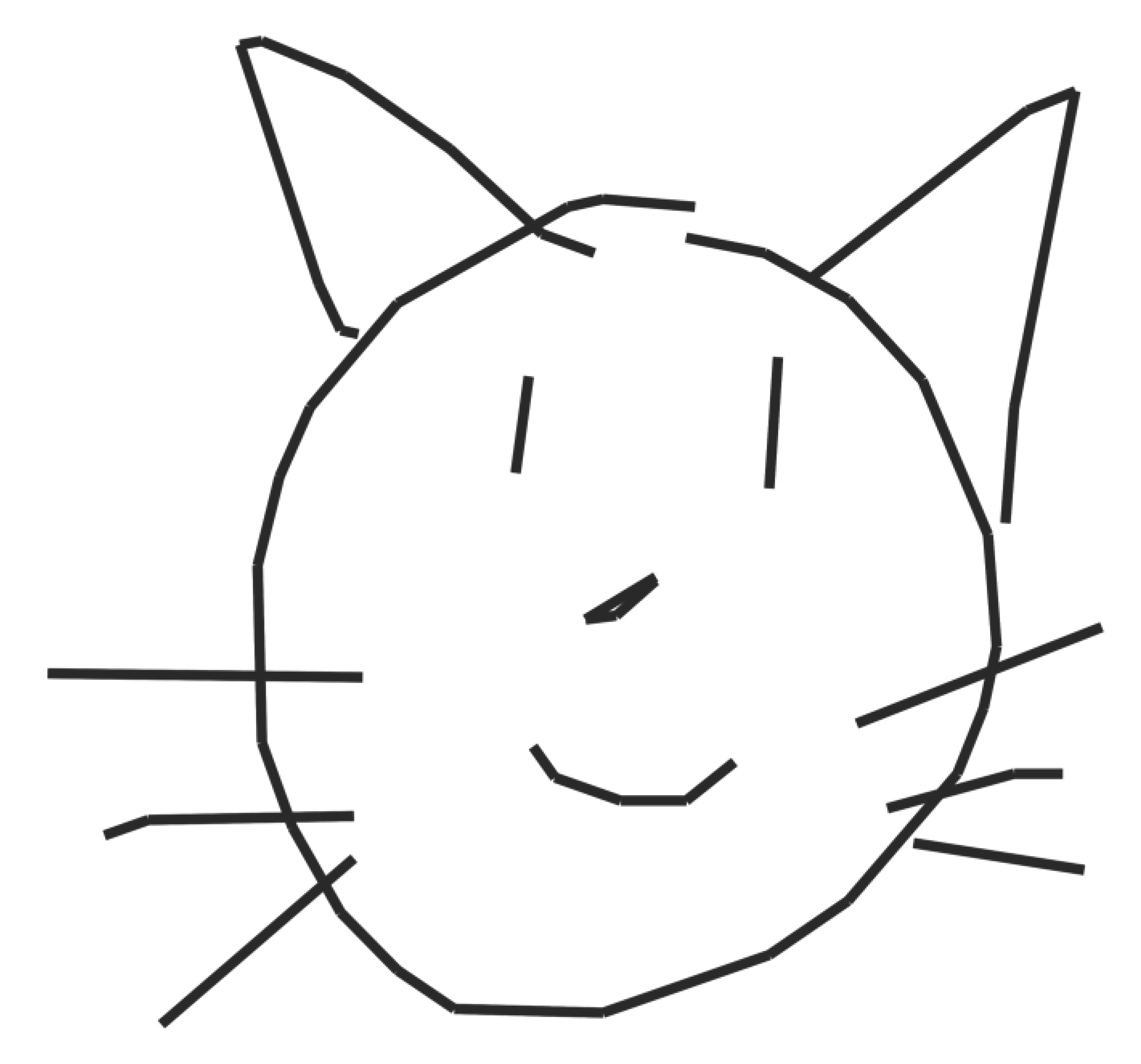}}
	%\hspace{5mm}	
	\subfigure[1\textrm{-hop} connected]{
		\label{fig:cat_3_b}
		\includegraphics[width=0.31\columnwidth]{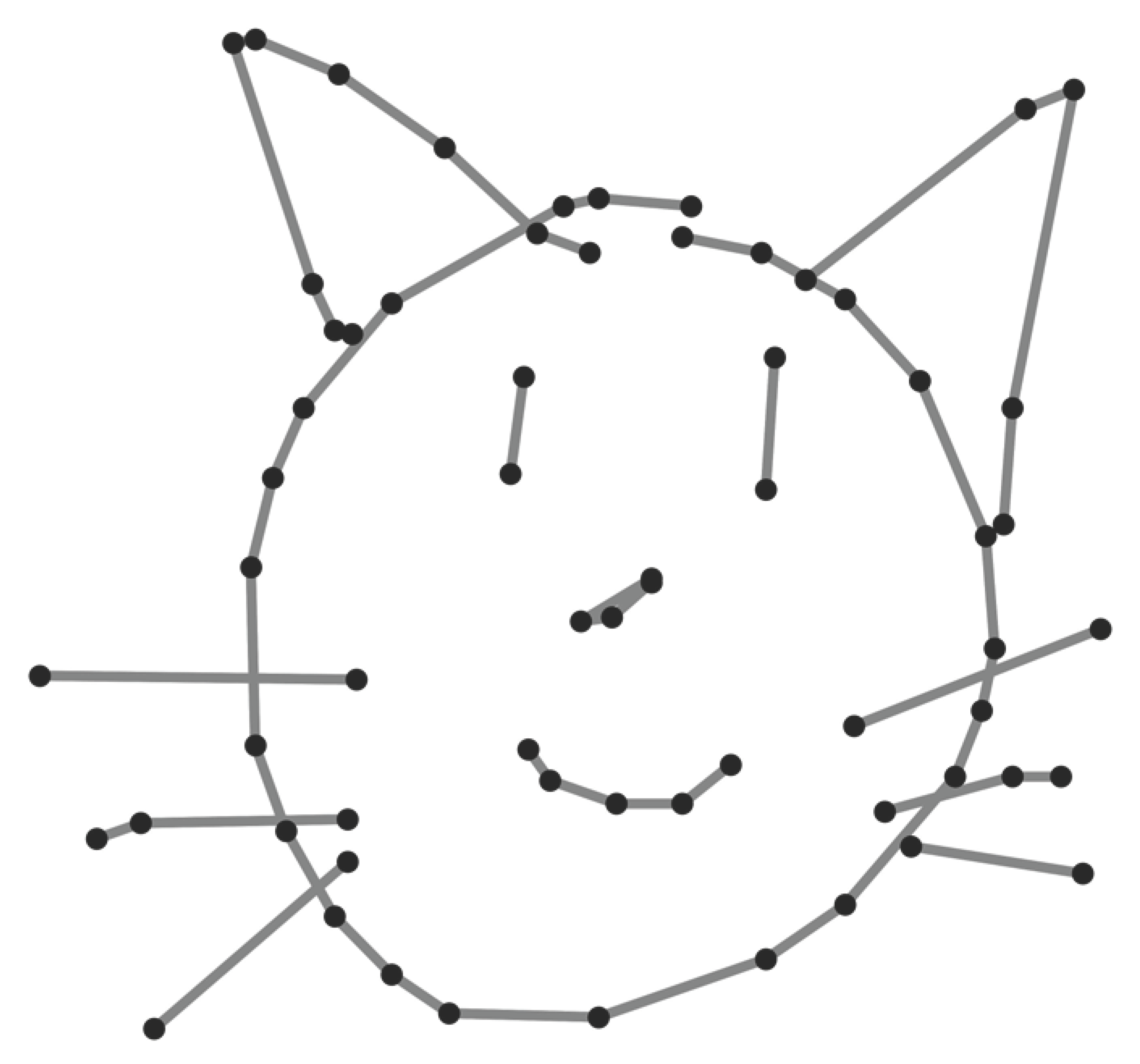}}
    \subfigure[2\textrm{-hop} connected]{
		\label{fig:cat_3_c}
		\includegraphics[width=0.31\columnwidth]{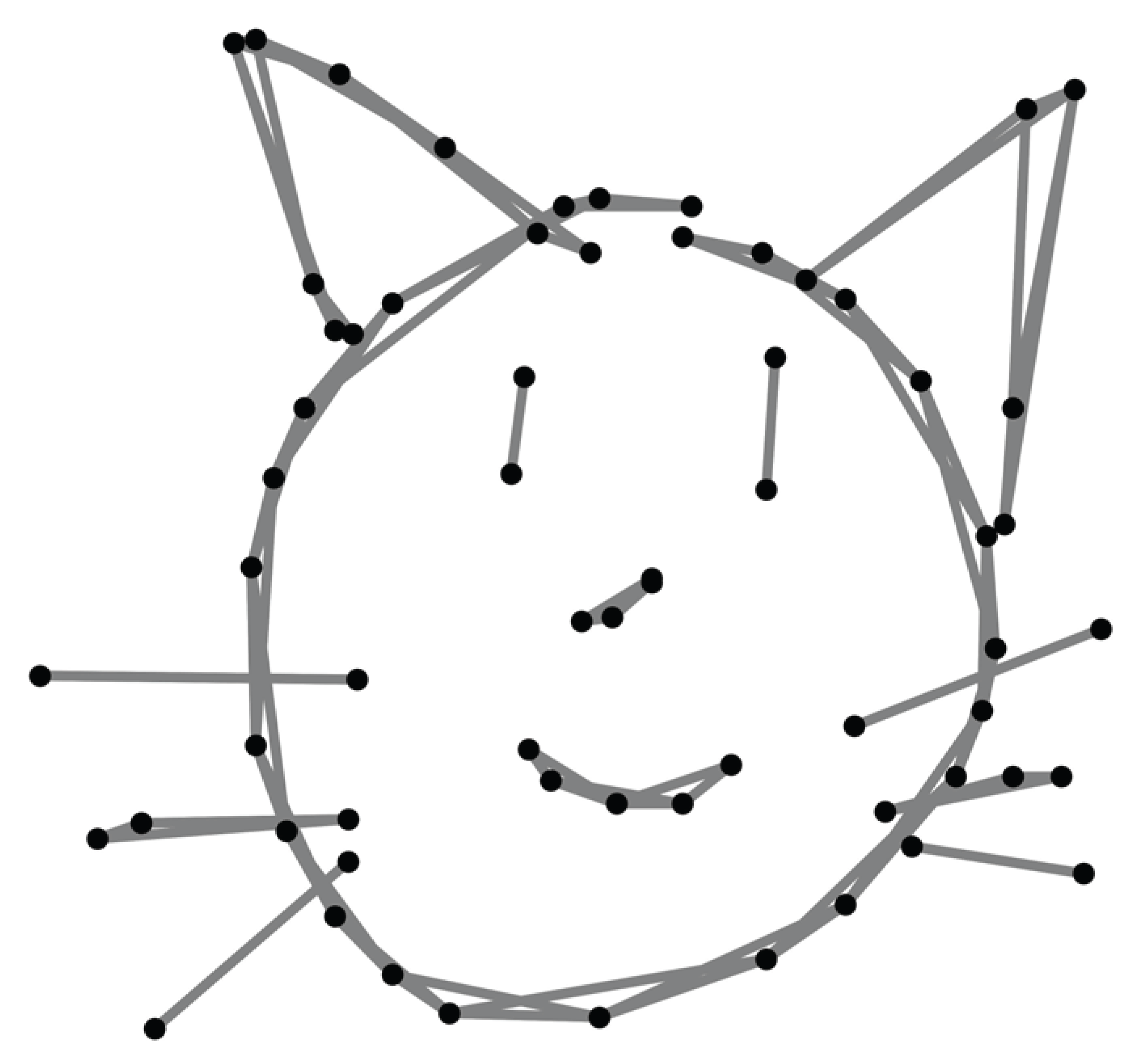}}
	\caption{Sketches can be seen as sets of curves and strokes, which are discretized by graphs.}
	\label{fig:motivation_1}
\end{figure}
%---------------------------------------------------------

\begin{figure}[!t]
\begin{center}
%\fbox{\rule{0pt}{2in} \rule{.9\linewidth}{0pt}}
\includegraphics[width=1.05\columnwidth]{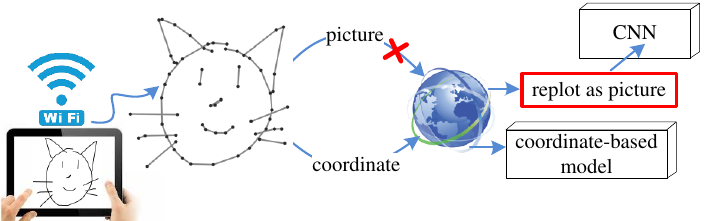}
\end{center}
%\vspace{-0.3cm}
   \caption{{In sketch-based human-computer interaction scenarios, it is time-consuming to render and transfer pictures of sketches. Solely transferring stroke coordinates leads to real-time applications.}}
   %\vspace{-0.5cm}
\label{fig:motivation_2}
\end{figure}

If we assume sketches to be 2D static images, CNNs can be directly applied to sketches, such as ``Sketch-a-Net''~\cite{yu2015sketch}. If we now suppose that sketches are ordered sequences of point coordinates, then RNNs can be used to recursively capture the temporal information, \eg, ``SketchRNN''~\cite{ha2017sketchrnn}.  

In this work, we introduce a new representation of sketches with {\it graphs}. We assume that sketches are sets of curves and strokes, which are discretized by a set of points representing the graph nodes. This view offers high flexibility to encode different sketch geometric properties as we can decide different connectivity structures between the node points. We use two types of graphs to represent sketches: intra-stroke graphs and extra-stroke graphs. The first graphs capture the local geometry of strokes, independently to each other, with for example 1\textrm{-hop} or 2\textrm{-hop} connected graphs, see Figure~\ref{fig:motivation_1}. The second graphs encode the global geometry and temporal information of strokes.   Another advantage of using graphs is the freedom to choose the node features. For sketches, spatial, temporal and semantic information is available with the stroke point coordinates, the ordering of points, and the pen state information, respectively. In summary, representing sketches with graphs offers a universal representation that can make use of global and local spatial sketch structures, as well as temporal and semantic information.

To exploit these graph structures, we propose a new Transformer \cite{vaswani2017attention} architecture that can use multiple sparsely connected graphs.
It is worth reporting that a direct application of the original Transformer model on the input spatio-temporal features provides poor results. 
We argue that the issue comes from the graph structure in the original Transformer which is a fully connected graph. 
Although fully-connected word graphs work impressively for Natural Language Processing, where the underlying word representations themselves contain rich information, such dense graph structures provide poor innate priors/inductive bias \cite{battaglia2018relational} for 2D sketch tasks.
Transformers require sketch-specific design coming from geometric structures. This led us to naturally extend Transformers to multiple arbitrary graph structures. Moreover, graphs provide more robustness to handle noisy and style-changing sketches as they focus on the geometry of stokes and not on the specific distribution of points.

Another advantage of using domain-specific graphs is to leverage the sparsity property of discretized sketches. Observe that intra-stroke and extra-stroke graphs are {\it highly sparse} adjacency matrices. In practical sketch-based human-computer interaction scenarios, it is time-consuming to directly transfer the original sketch picture from user touch-screen devices to the back-end servers. To ensure real-time applications, transferring the stroke coordinates as a character string would be more beneficial, see Figure~\ref{fig:motivation_2}.

Our {\textbf{main contributions}} can be summarised as follows: 
 {
\begin{enumerate}
\item We propose to model sketches as sparsely connected graphs, which are flexible to encode local and global geometric sketch structures. To the best of our knowledge, it is the first time that graphs are proposed for representing sketches.
\item We introduce a novel Transformer architecture that can handle multiple arbitrary graphs. Using intra-stroke and extra-stroke graphs, the proposed {\it Multi-Graph Transformer} (MGT) learns both local and global patterns along sub-components of sketches.
\item Numerical experiments demonstrate the performances of our model. MGT significantly outperforms RNN-based models, and
achieves small recognition gap to CNN-based architectures.
Moreover, our MGT infers faster than the strongest CNN baseline (see details in Section~\ref{sec:timecost}).
This is promising for real-time sketch-based human-computer interaction systems. 
Note that for sketch recognition, CNNs are the performance upper bound of coordinate-based models that involve truncating coordinate sequences, \eg, RNN or Transformer based architectures. 
\item This Multi-Graph Transformer model is agnostic to graph domains, and can be used beyond sketch applications. We transfer our multi-graph transformer idea to conduct Relation Extraction on a NLP benchmark, \ie, SemEval-2010 task 8, outperforming the state-of-the-art CNNs by a clear margin (prediction acc.: ours ($89.45\%$) vs. CNN ($86.60\%$)).
\end{enumerate}    }
%(i) We propose to model sketches as sparsely connected graphs, which are flexible to encode local and global geometric sketch structures. To the best of our knowledge, it is the first time that graphs are proposed for representing sketches.\\
%(ii) We introduce a novel Transformer architecture that can handle multiple arbitrary graphs. Using intra-stroke and extra-stroke graphs, the proposed {\it Multi-Graph Transformer} (MGT) learns both local and global patterns along sub-components of sketches. \\
%%(iii) This Multi-Graph Transformer model is agnostic to graph domains, and can be used beyond sketch applications.\\
%\black{
%(iii) Numerical experiments demonstrate the performances of our model. MGT significantly outperforms RNN-based models, and
%achieves small recognition gap to CNN-based architectures.
%This is promising for real-time sketch-based human-computer interaction systems. 
%Note that for sketch recognition, CNNs are the performance upper bound of coordinate-based models that involve truncating coordinate sequences, \eg, RNN or Transformer based architectures. 
%}\\
%(iv) This Multi-Graph Transformer model is agnostic to graph domains, and can be used beyond sketch applications. We transfer our multi-graph transformer idea to conduct Relation Extraction on a NLP benchmark, \ie, SemEval-2010 task 8, outperforming the state-of-the-art CNNs by a clear margin (prediction acc.: ours ($89.45\%$) vs. CNN ($86.60\%$)).

From the perspective of representing sketches and the practical sketch-oriented applications, the {\textbf{main advantages}} of our proposed Multi-Graph Transformer against CNN are:
{
\begin{enumerate}
\item Our model can be used for real-time sketch-based HCI~(Human Computer Interaction) systems, because it does not need to render and transfer sketch pictures (see Figure~\ref{fig:motivation_2}). 
\item Our MGT is faster than the strongest CNN baseline, \ie, Inception~V3. This will be demonstrated by experiments in Section~\ref{sec:timecost}.
\item As demonstrated in~\cite{li2019toward}, coordinate-based models (\eg, our MGT, RNNs) are more suitable than CNNs for the stroke-level tasks, \eg, perceptual grouping, stroke-grained segmentation.
\item  Our MGT takes the stroke sequence as input, thus it can handle the tasks that need to understand the logic and timing patterns of sketching process, \eg, stroke-by-stroke generation, stroke-level abstraction. 
Specifically, stroke-by-stroke sketch generation needs models learn the temporal information, because the models need to decide the order of stroke generation.
However, as discussed in~\cite{xu2020deep}, CNNs intrinsically fail to work for these tasks.
\end{enumerate}}

 The rest of this paper is organized as follows: Section~\ref{sec:related_work}
 briefly summarizes related work. Section~\ref{sec:methodology} describes our
 proposed Multi-Graph Transformer.
 Experimental results and discussion are presented
 in Section~\ref{sec:experiments}, followed by a conclusion Section~\ref{sec:conclusion}. 
\black{Finally, we discuss our future work in Section~\ref{sec:future_work}}.

%-------------------------------------------------------------------------------------
\section{Related Work}
\label{sec:related_work}

%\noindent\textbf{Neural Network Architectures for Sketches}
\subsection{Neural Network Architectures for Sketches } 
% \paragraph{Neural Network Architectures for Sketches}
CNNs are a common choice for feature extraction from sketches. ``Sketch-a-Net''~\cite{yu2015sketch} is a representative CNN-based model having a sketch-specific architecture. 
It was directly inspired from AlexNet~\cite{krizhevsky2012imagenet} with larger first layer filters, no layer normalization, larger pooling sizes, and high dropout. Song~\etal~\cite{song2017deep} further improved Sketch-a-Net by adding spatial-semantic attention layers. ``SketchRNN''~\cite{ha2017sketchrnn} is a seminal work to model temporal stroke sequences with RNNs, {by taking the key point coordinates of stroke as input. ``SketchRNN'' reminds the researchers that sketching is a dynamic process so that the temporal patterns of sketching strokes should also be considered.} 
%A CNN-RNN hybrid architecture for sketches was proposed in \cite{sarvadevabhatla2016enabling}.

%\blue{Recently, some CNN-RNN hybrid architectures have been proposed for sketches, \eg, CNN-RNN dual-branch network~\cite{xu2018sketchmate}, CNN-to-RNN cascaded network~\cite{sarvadevabhatla2016enabling}, RNN-RNN dual-branch network~\cite{jia2017sequential}, RNN-to-CNN cascaded network~\cite{9068451}. For more detailed summary and comparison for sketch representing networks, please check a comprehensive survey~\cite{xu2020deep}.}

{Recently, some CNN-RNN hybrid architectures have been proposed for sketches, \eg, dual-branch networks~\cite{jia2017sequential,xu2018sketchmate}, cascaded networks~\cite{sarvadevabhatla2016enabling,9068451}. For more detailed summary and comparison for sketch representing networks, please check a comprehensive survey~\cite{xu2020deep}.}

{Essentially, the aforementioned CNN or RNN -based network architectures model sketch in  Euclidean space. How to model sketch in  non-Euclidean spaces is an interesting question.}
In this work, we propose a novel Graph Neural Network architecture for learning sketch representations from multiple sparse graphs, combining both stroke geometry and temporal order.

%\vspace{3pt}
%\noindent\textbf{Graph Neural Networks }
\subsection{Graph Neural Networks }
% \paragraph{Graph Neural Networks}
Graph Neural Networks (GNNs) \cite{wu2020comprehensive,bruna2014spectral,defferrard2016convolutional,sukhbaatar2016gcn,kipf2017semi,hamilton2017inductive,monti2017geometric,shanthamallu2019gramme} aim to generalize neural networks to non-Euclidean domains such as graphs and manifolds.
GNNs iteratively build representations of graphs through recursive neighborhood aggregation (or message passing),
where each graph node gathers features from its neighbors to represent local graph structure.
{Recently, Graph Neural Networks have been widely applied for various domains, \eg, vision, language. However, sketch-oriented Graph Neural Networks are still under-studied.}

%\vspace{3pt}
%\noindent\textbf{Transformers }
\subsection{Transformers}
% \paragraph{Transformers}
The Transformer architecture \cite{vaswani2017attention}, originally proposed as a powerful and scalable alternative to RNNs, has been widely adopted in the Natural Language Processing community for tasks such as machine translation \cite{edunov2018understanding,wang2019learning}, language modelling \cite{radford2018improving,dai2019transformer}, and question-answering \cite{devlin2019bert,yang2019xlnet}.

Transformers for NLP can be regarded as GNNs which use self-attention \cite{bahdanau2014neural,velickovic2018graph} for neighborhood aggregation on fully-connected word graphs \cite{ye2019bp}.
{However, GNNs and Transformers perform poorly when sketches are modelled as fully-connected graphs.
This work advocates for the injection of inductive bias into Transformers through domain-specific graph structures.}

%-------------------------------------------------------------------------

\section{Method}
\label{sec:methodology}

\subsection{Notation} We assume that the training dataset $D$ consists of $N$ labeled sketches: $D = \{({\bf X}_{n}, z_{n})\}_{n = 1}^{N}$. 
%$y_{n}$ is the class label of $n$-th sketch $S_{n}$.
Each sketch ${\bf X}_{n}$ has a class label $z_{n}$, and can be formulated as a $S$-step sequence $[{\bf C}_{n}, {\bf f}_{n}, {\bf p}] \in \mathbb{R}^{S \times 4}$.
${\bf C}_n = \{(x_{n}^{s}, y_{n}^{s})\}_{s = 1}^{S} \in \mathbb{R}^{S \times 2}$ is the coordinate sequence of the sketch points ${\bf X}_n$. 
All sketch point coordinates have been uniformly scaled to $x_{n}^{s},y_{n}^{s} \in [0,~256]^2$.
If the true length of ${\bf C}_n$ is shorter than $S$ then the vector $[-1, -1]$ is used for padding.  
Flag bit vector ${\bf f}_n \in \{f_{1}, f_{2}, f_{3}\}^{S \times 1}$ is a ternary integer vector that denotes the pen state sequence corresponding to each point of ${\bf X}_n$. It is defined as follows:
$f_{1}$ if the point $(x_{n}^{s}, y_{n}^{s})$  is a starting or ongoing point of a stroke, $f_{2}$ if the point is the ending point of a stroke, and $f_{3}$ for a padding point.
%Thus, the true length $t_{n}$ of ${\bf X}_n$ can be easily deduced by ${\bf C}_n$ or ${\bf f}_n$. 
Vector ${\bf p}=[0, 1, 2, \cdots, S - 1]^{T}$  is a positional encoding vector that represents the temporal position of the points in each sketch ${\bf X}_{n}$.

Given $D$, we aim to model ${\bf X}_n$ as multiple sparsely connected graphs and learn a deep embedding space, where the high-level semantic tasks can be conducted, \eg, sketch recognition.

\begin{figure}[!t]
    \centering
    \includegraphics[width=1.07\columnwidth]{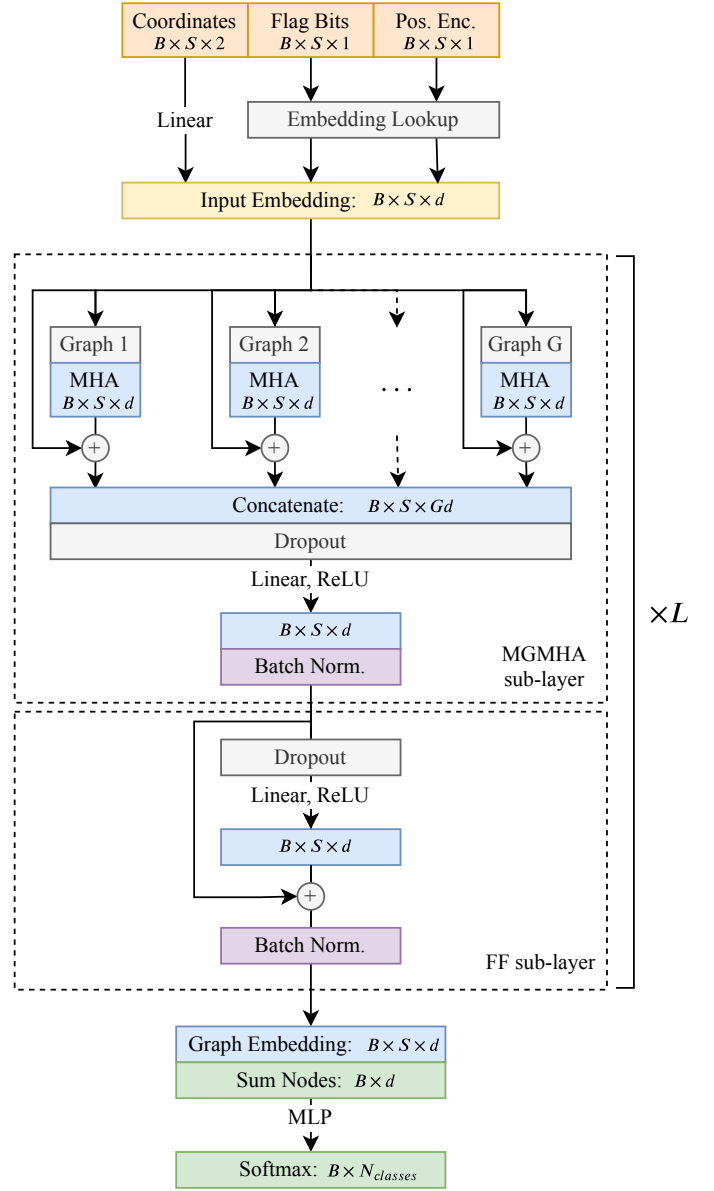}
    \caption{Multi-Graph Transformer architecture.
    Each MGT  layer is composed of (i) a Multi-Graph Multi-Head  Attention (MGMHA) sub-layer and (ii) a position-wise fully  connected Feed-Forward (FF) sub-layer. See details in text.
    ``B'' denotes batch size.}
    \label{fig:pipeline_details}
\end{figure}

\subsection{Multi-Modal Input Layer}
\label{input-layer}
Given a sketch ${\bf X}_n$, we model its $S$ stroke points as  $S$ nodes of a graph. Each node has three features: (i) ${\bf C}_{n}^{s}$ is the spatial positional information of the current stroke point $s$, (ii) ${\bf f}_{n}^{s}$ is the pen state of the current stroke point. This information helps to identify the stroke points belonging to the same stroke, 
and (iii) ${\bf p}^{s}$ is the temporal  information of the current stroke point. As sketching is a dynamic process, it is important to use the temporal information.

The complete model architecture for our Multi-Graph Transformer is presented in Figure \ref{fig:pipeline_details}.
Let us start by describing the input layer. The final vector at node $s$ of the multi-modal input layer is defined as
\begin{equation}
\label{equ:input}
({\bf h}_{n}^{s})^{(l=0)} = \mathcal{C}( \mathcal{E}_1( {{\bf C}_{n}^{s}}), \mathcal{E}_2({\bf f}_{n}^{s}), \mathcal{E}_2({\bf p}^{s})),
\end{equation}
where $ \mathcal{E}_1( {{\bf C}_{n}^{s}})$ is the embedding of ${\bf C}_{n}^{s}$ with a linear layer of size $2\times \hat{d}$, $\mathcal{E}_2({\bf f}_{n}^{s})$ and $\mathcal{E}_2({\bf p}^{s})$ are the embeddings of the flag bit ${\bf f}_{n}^{s}$ (3 discrete values) and the position encoding ${\bf p}^{s}$ ($S$ discrete values) from an embedding dictionary of size $(S + 3)\times \hat{d}$,
and $\mathcal{C}(\cdot, \cdot)$ is the concatenation operator. The node vector $({\bf h}_{n}^{s})^{(l=0)}$ has dimension $d = 3 \hat{d}$. 
The design of the input layer was selected after extensive ablation studies, which are described in subsequent sections. 

%-------------------------------------------------------

\begin{figure}[!t]
    \centering
    \includegraphics[width=1.1\columnwidth]{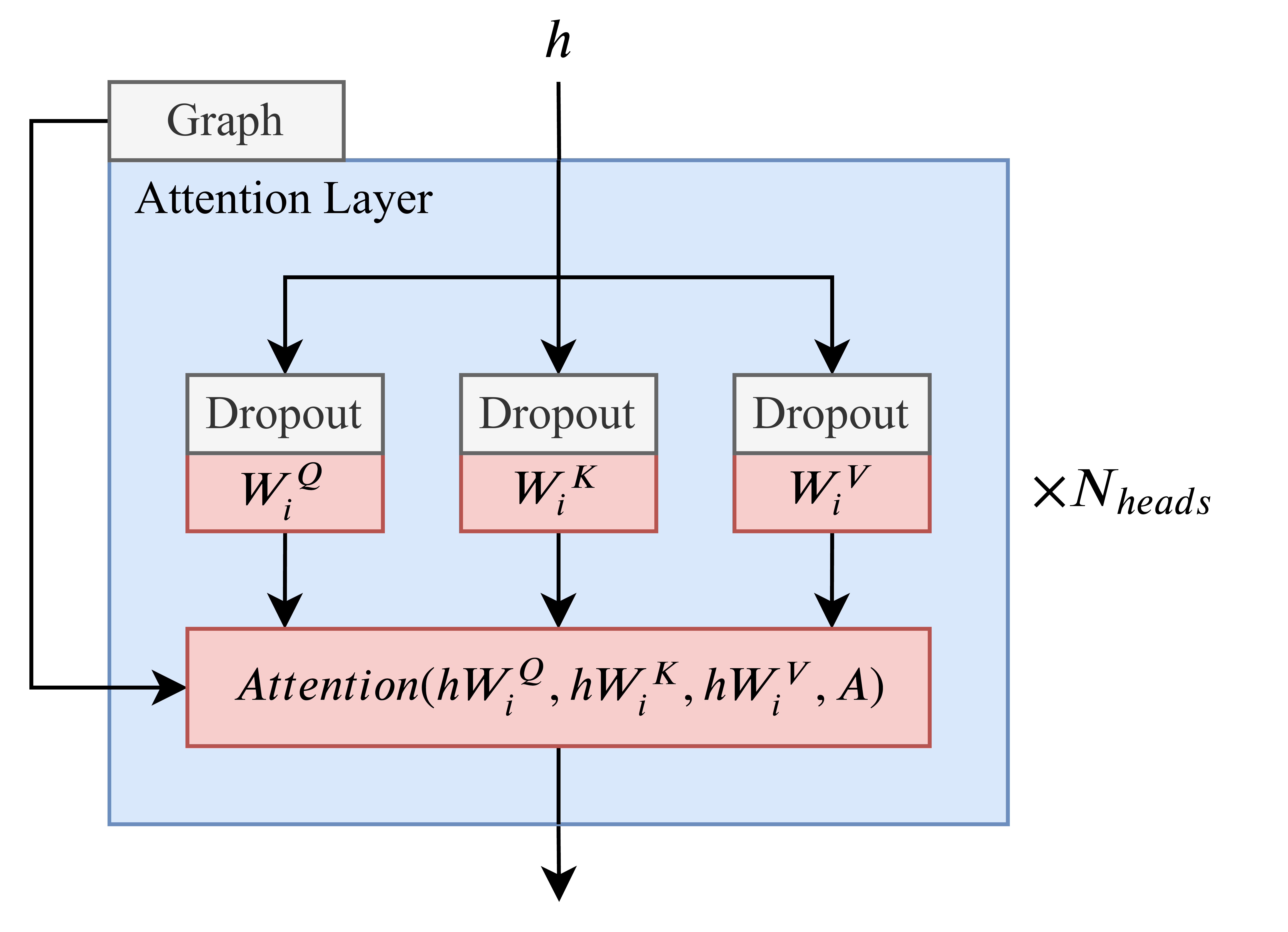}
    \caption{Multi-Head Attention Layer, consisting of several Graph Attention Layers in parallel.}
    \label{fig:attention}
\end{figure}

\subsection{Multi-Graph Transformer}
\label{sec:MGT}

The initial node embedding $({\bf h}_{n}^{s})^{(l=0)}$ is updated by stacking $L$ Multi-Graph Transformer (MGT) layers \eqref{equ:mgt}. Let us describe all layers.\\

\subsubsection{Graph Attention Layer} 
%\noindent\textbf{Graph Attention Layer } 
Let ${\bf A}$ be a graph adjacency matrix of size $S \times S$ and ${\bf Q} \in \mathbb{R}^{S \times d_{q}}, {\bf K} \in \mathbb{R}^{S \times d_{k}}, {\bf V} \in \mathbb{R}^{S \times d_{v}}$ be the query, key, and value matrices. We define a graph attention layer as
\begin{equation}
\label{equ:attention}
\textrm{GraphAttention}({\bf Q}, {\bf K}, {\bf V}, {\bf A}) = {\bf A} \ \odot \ \textrm{softmax} ( \frac{{\bf Q}{\bf K}^{T}}{\sqrt{d_k}} ) {\bf V}, 
\end{equation}
where $\odot$ is the Hadamard product. We simply weight the ``Scaled Dot-Product Attention''~\cite{vaswani2017attention} with the graph edge weights. 
We set $d_{q} = d_{k} = d_{v} = \frac{d}{I}$, where $I$ is the number of attention heads.
\\

\subsubsection{Multi-Head Attention Layer} 
%\noindent\textbf{Multi-Head Attention Layer } 
We aggregate the graph attentions with multiple heads:
\begin{equation}
\label{equ:mha}
\textrm{MultiHead}({\bf Q},{\bf K},{\bf V},{\bf A}) = \mathcal{C} ( \textrm{head}_{1}, \!\cdots\!, \textrm{head}_{I} ) {\bf W}^{O}, 
\end{equation}
where ${\bf W}^{O} \in \mathbb{R}^{Id_v \times d}$
and each attention head is computed with the graph attention layer \eqref{equ:attention}:
\begin{equation}
\label{equ:mha-head}
\textrm{head}_{i} = \textrm{GraphAttention}({\bf Q}{\bf W}_{i}^{Q}, {\bf K}{\bf W}_{i}^{K}, {\bf V}{\bf W}_{i}^{V}, {\bf A}), 
\end{equation}
where ${\bf W}_{i}^{Q} \in \mathbb{R}^{d \times d_q}$, ${\bf W}_{i}^{K} \in \mathbb{R}^{d \times d_k}$, and ${\bf W}_{i}^{V} \in \mathbb{R}^{d \times d_v}$.
We add dropout \cite{srivastava2014dropout} before the linear projections of $\bf Q$, $\bf K$ and $\bf V$.
An illustration of the Multi-Head Attention Layer is presented in Figure \ref{fig:attention}.
\\

\subsubsection{Multi-Graph Multi-Head Attention Layer }
%\noindent\textbf{Multi-Graph Multi-Head Attention Layer } 
Given a set of adjacency graph matrices
$\{{\bf A}_{g}\}_{g=1}^{G}$, we can concatenate Multi-Head Attention Layers:
\begin{equation}
\label{equ:mgmha}
\begin{split}
&\textrm{MultiGraphMultiHeadAttention}({\bf Q},{\bf K},{\bf V},\{{\bf A}_{g}\}_{g=1}^{G}) = \\  & \textrm{ReLU}(\mathcal{C} ( \textrm{ghead}_{1},\cdots, \textrm{ghead}_{G} ) {\bf W}^{\widetilde{O}}),
\end{split}
\end{equation}
where ${\bf W}^{\widetilde{O}}\in\mathbb{R}^{Gd \times d}$ and each Multi-Head Attention Layer is computed with \eqref{equ:mha}:
\begin{equation}
\label{equ:mgmha-head}
\textrm{ghead}_{g} = \textrm{MultiHead}({\bf Q},{\bf K},{\bf V},{\bf A}_{g}).
\end{equation}

\subsubsection{Multi-Graph Transformer Layer} 
%\noindent\textbf{Multi-Graph Transformer Layer } 
The Multi-Graph Transformer (MGT) at layer $l$ for node $s$ is defined as
% \begin{eqnarray}
% \label{equ:kvds}
% ({\bf h}_{n}^{s})^{(l)} &=& \textrm{MGT}(({\bf h}_{n})^{(l-1)}),\\
% &=& \textrm{BN}^{(l)} (\hat{{\bf h}}_{n}^{s} + \textrm{FF}^{(l)} (\hat{{\bf h}}_{n}^{s})),
% \end{eqnarray}
\begin{equation}
\label{equ:mgt}
\begin{split}
({\bf h}_{n}^{s})^{(l)} &= \textrm{MGT}(({\bf h}_{n})^{(l-1)}) \\ 
&=\hat{{\bf h}}_{n}^{s} + \textrm{FF}^{(l)} (\hat{{\bf h}}_{n}^{s}),
\end{split}
\end{equation}
where the intermediate feature representation $\hat{{\bf h}}_{n}^{s}$ is defined as:
\begin{equation}
\label{equ:mgt-int}
\hat{{\bf h}}_{n}^{s} = (\textrm{MGMHA}_{n}^{s})^{(l)} (({\bf h}_{n}^{1})^{(l - 1)}, \cdots, ({\bf h}_{n}^{S})^{(l - 1)}).
\end{equation}
The MGT layer is thus composed of (i) a Multi-Graph Multi-Head Attention~(MGMHA) sub-layer \eqref{equ:mgmha} and (ii) a position-wise fully connected Feed-Forward (FF) sub-layer.
Each MHA sub-layer \eqref{equ:mgmha-head} and FF \eqref{equ:mgt} has residual-connection~\cite{he2016deep} and batch normalization~\cite{Ioffe2015}.
See Figure~\ref{fig:pipeline_details} for an illustration.

\subsection{Sketch Embedding and Classification Layer}
Given a sketch ${\bf X}_n$ with $t_n$ key points, its continuous representation ${\bf h}_{n}$ is simply given by the sum over all its node features from the last MGT layer:
\begin{equation}
\label{equ:kjoa}
{\bf h}_{n} = \sum_{s=1}^{t_{n}} ({\bf h}_{n}^{s})^{(L)}.
\end{equation}
Finally, we use a Multi-Layer Perceptron (MLP) to classify the sketch representation ${\bf h}_{n}$, see Figure~\ref{fig:pipeline_details}.

\subsection{Sketch-Specific Graphs}
In this section, we discuss the graph structures we used in our Graph Transformer layers. We  considered two types of graphs, which capture local and global geometric sketch structures. 

The first class of graphs focus on representing the local geometry of individual strokes. We choose $K$-hop graphs to describe the local geometry of strokes.  The intra-stroke adjacency matrix is defined as follows:
\begin{equation}
\label{equ:attention_mask_k_hops}
{\bf A}_{n,ij}^{K\textrm{-hop}} = \left\{
\begin{aligned}
1 & ~~\textrm{ if } j \in \mathcal{N}_i^{K\textrm{-hop}} \textrm{ and } j \in \textrm{\textrm{global}}(i),\\
0 & ~~\textrm{ otherwise },
\end{aligned}
\right.
\end{equation}
where $\mathcal{N}_i^{K\textrm{-hop}}$ is the K-hop neighborhood of node $i$ and $ \textrm{\textrm{global}}(i)$ is the stroke of node $i$.

The second class of graphs capture the global and temporal relationships between the strokes composing the whole sketch. We define the extra-stroke adjacency matrix as follows:
\begin{equation}
\label{equ:attention_mask_stroke_level}
{\bf A}_{n,ij}^{\textrm{global}} =\left\{
\begin{aligned}
1 & ~~\textrm{ if } |i-j|= 1 \textrm{ and } \textrm{\textrm{global}}(i) \not= \textrm{\textrm{global}}(j),\\
0 & ~~\textrm{ otherwise }.
\end{aligned}
\right.
\end{equation}
This graph will force the network to pay attention between two points belonging to two distinct strokes but consecutive in time,
thus allowing the model to understand the relative arrangement of strokes.

%------------------------------------------------------
\section{Experiments}
\label{sec:experiments}

\subsection{Experimental Setting} 
\label{expt-settings} 
In this section, we detail our experimental settings. \\

\subsubsection{Dataset and Pre-Processing} 
\begin{figure}[!t]
	\centering
	%\hspace{5mm}	
	\subfigure[Training]{
		\label{fig:train_set_hist}
      \includegraphics[width=0.23\textwidth]{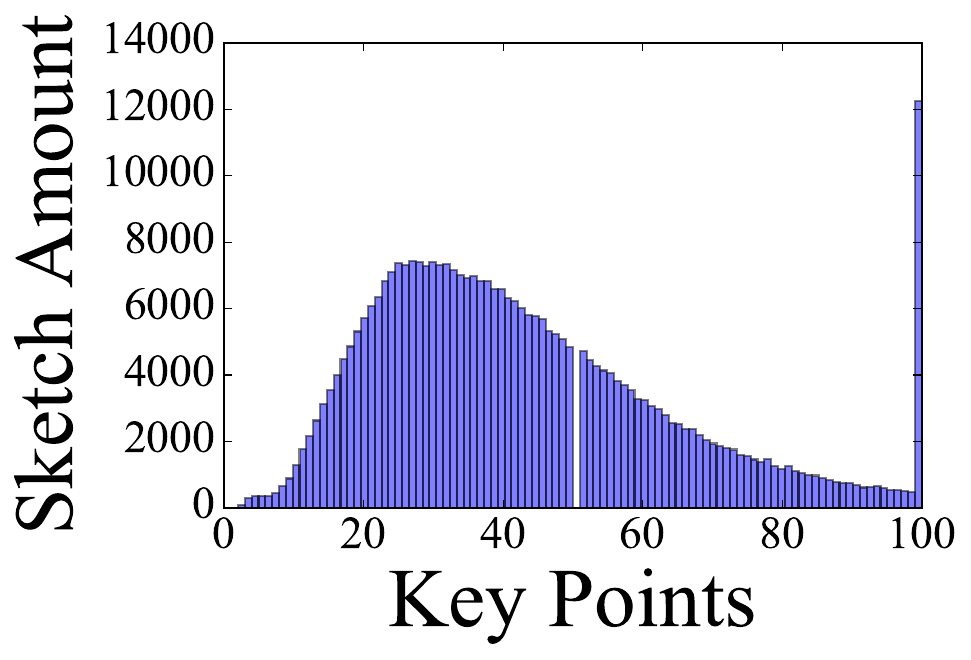}}
    %\hspace{5mm}
    \subfigure[Validation]{
		\label{fig:val_set_hist}
      \includegraphics[width=0.23\textwidth]{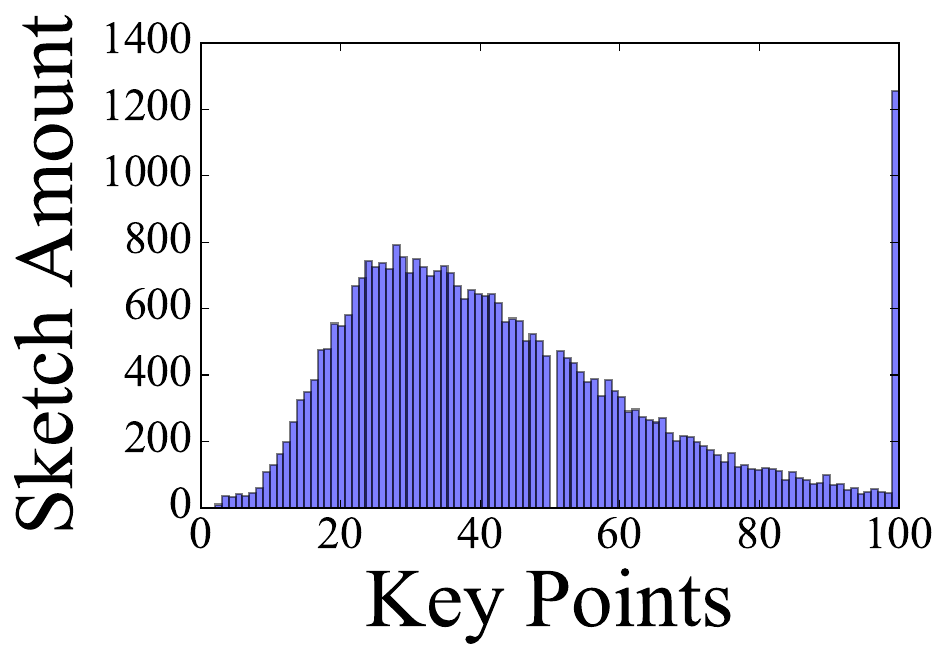}}
    %\hspace{5mm}
    \subfigure[Test]{
		\label{fig:test_set_hist}
      \includegraphics[width=0.23\textwidth]{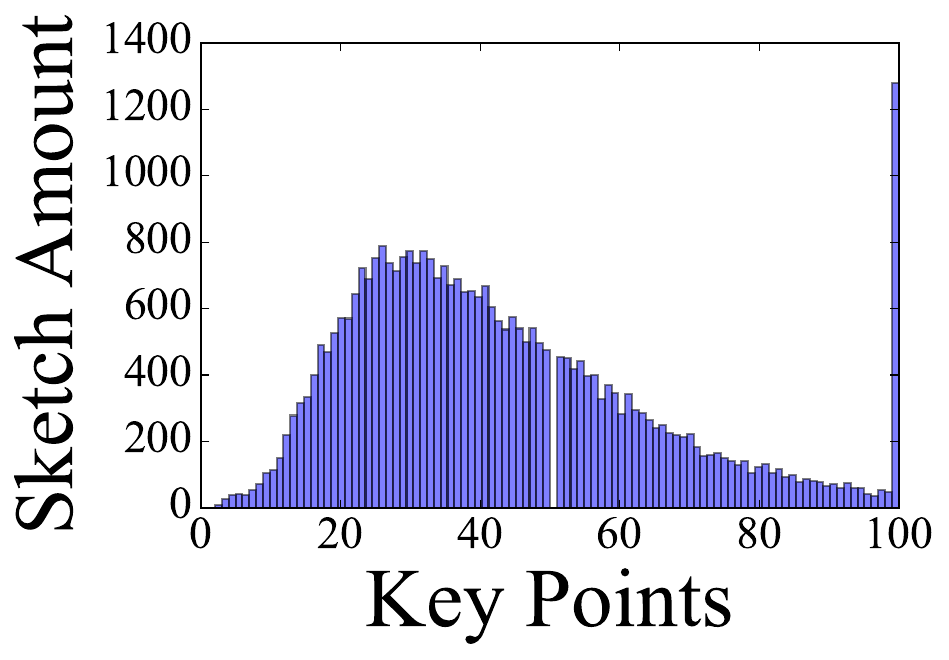}}
    %\hspace{5mm}
	\caption{Histograms of key points per sketch for our subset of QuickDraw. The sharp spike at $100$ key points is due to truncation.}
	%\vspace{-0.6cm}
	\label{fig:statistic_analysis_4_dataset}
\end{figure}

%\noindent\textbf{Dataset and Pre-Processing } 
% \paragraph{Dataset and Pre-processing}
Google QuickDraw~\cite{ha2017sketchrnn}~\footnote{\url{https://quickdraw.withgoogle.com/data}}~\footnote{\url{https://github.com/googlecreativelab/quickdraw-dataset}} is the largest available sketch dataset containing 50 Million sketches as simplified stroke key points in temporal order, sampled using the Ramer–Douglas–Peucker algorithm after uniformly scaling image coordinates within $0$ to $256$.
Unlike smaller crowd-sourced sketch datasets, \eg, TU-Berlin~\cite{eitz2012humans}, QuickDraw samples were collected via an international online game where users have only 20 seconds to sketch objects from 345 classes, such as cats, dogs, clocks, \textit{etc}.
Thus, sketch classification on QuickDraw not only involves a diversity of drawing styles, but can also be highly abstract and noisy, making it a challenging and practical test-bed for comparing the effectiveness of various neural network architectures.
Following recent practices \cite{Dey_2019_CVPR,xu2018sketchmate}, we create random training, validation and test sets from the full dataset by sampling $1000$, $100$ and $100$ sketches respectively from each of the 345 categories in QuickDraw.
\black{Although the transformer architecture is able to handle sketches with any finite number of key points,}
following \cite{xu2018sketchmate}, we truncate or pad all samples to a uniform length of 100 key points/steps to facilitate efficient training of RNN and GNN-based models, \black{to save parameters and training time}. 
We provide summary statistics for our training, validation and test sets in Table~\ref{table:tiny_quickdraw_v1}, and histograms visualizing the key points per sketch are shown in Figure \ref{fig:statistic_analysis_4_dataset}. \\

%\vspace{3pt}
\subsubsection{Evaluation Metrics}
%\noindent\textbf{Evaluation Metrics}\quad
Our evaluation metric for sketch recognition is ``top K accuracy'', the proportion of samples whose true class is in the top K model predictions, for values  $k = 1, 5, 10$.
(Note that acc.@k $= 1.0$ means 100\%) \\

%------------------------------------------------------------------------
\begin{table}[!t]
\caption{Summary statistics for our subset of QuickDraw.}
\label{table:tiny_quickdraw_v1}
%\small
%\footnotesize
%\scriptsize
%\tiny
\begin{center}
\resizebox{\columnwidth}{!}{
\begin{tabular}{c || c | c | c | c | c | c }
\hline
\multirow{2}{*}{Set} & \multirow{2}{*}{\# Samples} & \multirow{2}{*}{\# Truncated (ratio)} & \multicolumn{4}{c}{\# Key Points} \\
\cline{4-7}
  &   &  &  max & min & mean & std \\
\hline
Training & 345,000 & 11788~(3.42\%) & 100 & 2 & 43.26 & 21.85 \\
Validation & 34,500 & 1218~(3.53\%) & 100 & 2 & 43.24 &  21.89 \\
Test & 34,500 & 1235~(3.58\%) & 100 & 2 & 43.20 & 21.93 \\
\hline
\end{tabular}
}
\end{center}
\end{table}

%-----------------------------------------------------------------------------

%\vspace{3pt}
\subsubsection{Implementation Details} 
%\noindent\textbf{Implementation Details } 
% \paragraph{Implementation Details}
For fair comparison under similar hardware conditions, all experiments were implemented in PyTorch \cite{paszke2019pytorch}~\footnote{\url{https://pytorch.org/}} and run on one Nvidia 1080Ti GPU.
For Transformer models, we use the following hyperparameter values:
$S = 100$, $L = 4$, $\hat{d} = 128$, $G = 3$ (${\bf A}^{1\textrm{-hop}},{\bf A}^{2\textrm{-hop}},{\bf A}^{\textrm{global}}$), and $I = 8$ (per graph) for our Base model (and $\hat{d} = 256$ for our Large model).
Our FF sub-layer is a $d$-dimensional linear layer ($d = 3\hat{d}$) followed by ReLU \cite{glorot2011deep} and dropout.
The MLP Classifier consists of two $4\hat{d}$-dimensional linear layers with ReLU and dropout, followed by a $345$-dimensional linear projection representing logits over the 345 categories in QuickDraw.
We train all models by minimizing the softmax cross-entropy loss using the Adam \cite{kingma2014adam} optimizer for $100$ epochs.
We use an initial learning rate of $5e\!-\!5$ and multiply by a factor $0.7$ every $10$ epochs.
We use an early stopping strategy \black{(with the hyper-parameter ``patience'' of 10 epochs)} for selecting the final model, and
the checkpoint with the highest validation performance is chosen to report test performance. \\

%\vspace{3pt}
\subsubsection{Baselines}
%\noindent\textbf{Baselines }
% \paragraph{Baselines}
(i) From the perspective of coordinate-based sketch recognition, RNN models are a simple-yet-effective baseline.
Following Xu~\etal~\cite{xu2018sketchmate}, we design several bi-directional LSTM \cite{hochreiter1997long} and GRU \cite{cho2014properties} models at increasing parameter budgets comparable with MGT.
The final RNN states are concatenated and passed to the MLP classifier described previously.
We use batch size $256$, initial learning rate $1e\!-\!4$ and multiply by $0.9$ every $10$ epochs.
We train models with both our multi-modal input (Section \ref{input-layer}) as well as the 4D input from~\cite{xu2018sketchmate}.

(ii) Although converting sketch coordinates to images adds time overhead in practical settings and can be seen as auxilary information, we compare MGT to various state-of-the-art CNN architectures.
It is important to note that sketch sequences were truncated/padded for training both MGT and RNNs, hence image-based CNNs stand as an upper bound in terms of performance.
\black{For Inception V3~\cite{szegedy2016rethinking} and MobileNet V2~\cite{Sandler_2018_CVPR}, initial learning rate is $1e\!-\!3$ and multiplied by $0.5$ every $10$ epochs.
For other CNN baselines, the initial learning rate and decay are configured following their original papers.
}
\black{
For each model, we use the maximum possible batch size. Following standard practice in computer vision \cite{he2016deep,huang2017densely}, we employ early stopping based on observing over-fitting in the validation loss, 
and select the checkpoint with the highest validation accuracy for evaluation on the test set.
}

(iii) To evaluate the effectiveness of the proposed Graph Transformer layer,
we compare it with popular GNN variants: the Graph Convolutional Network \cite{kipf2017semi} and the Graph Attention Network \cite{velickovic2018graph} \footnote{For GAT, we use the same scaled dot-product attention mechanism as GT for efficiency.}.
All GNN models follow the same hyperparameter setup as Transformers
($L = 4$, $\hat{d} = 256$)
and are augmented with residual connections and batch normalization for fair comparison, following \cite{bresson2018experimental}.
\black{
Optimal hyper-parameters and learning rate schedules are selected based on validation set performance.
}

%------------------------------------------------------------------------
\begin{table*}[!t]
\caption{
Test set performance of MGT vs. the state-of-the-art RNN and CNN architectures. The $1^{st}$/$2^{nd}$/$3^{rd}$~best results per column are indicated in \textcolor{red}{\textbf{red}}/\textcolor{blue}{blue}/\textcolor{magenta}{magenta}.
}

\label{table:comparison_with_cnn_rnn}
%\small
%\footnotesize
%\scriptsize
%\tiny
\begin{center}
\resizebox{\textwidth}{!}{
\begin{tabular}{l || l | c | c | c | r }
\hline
  \multirow{2}{*}{Network} & \multirow{2}{*}{Configurations} & \multicolumn{3}{c|}{Recognition Accuracy} & \multirow{2}{*}{\tabincell{c}{Parameter \\ Amount}}  \\
  \cline{3-5}
 & & acc.@1 & acc.@5 & acc.@10 &   \\
\hline
\hline
% PUT IN SUPPLEMENT--> Bidirectional-LSTM \#1 & 4D input, 4 layer LSTM (hid\_size 256, dropout 0.5) & 0.6201 & 0.8564 & 0.9052 & 5,444,441  \\
 Bi-directional LSTM  \#1 & \small{4D Input, $\hat{d}=256, L=4, Dropout_{LSTM}=0.5, Dropout_{MLP}=0.15$} & 0.6665 & 0.8820 & 0.9189 &  5,553,241  \\
% PUT IN SUPPLEMENT--> Bidirectional-LSTM \#3 & 4D input, 5 layer LSTM (hid\_size 256, dropout 0.5) & 0.6172 & 0.8527 & 0.9025 & 7,021,401  \\
 Bi-directional LSTM  \#2 & \small{4D Input, $\hat{d}=256, L=5, Dropout_{LSTM}=0.5, Dropout_{MLP}=0.15$} & 0.6524 & 0.8697 & 0.9133 &  7,130,201  \\
 % PUT IN SUPPLEMENT-->  Bidirectional-GRU \#1 & 4D input, 5 layer GRU (hid\_size 256, dropout 0.5) & 0.6677 & 0.8845 & 0.9236 & 5,310,297  \\
 Bi-directional GRU & \small{4D Input, $\hat{d}=256, L=5, Dropout_{GRU}=0.5, Dropout_{MLP}=0.15$} & 0.6768 & 0.8854 & 0.9234 & 5,419,097  \\
 %Bi-LSTM                 & 4D input & poor & - & - & -  \\
 \hline
 \hline
 AlexNet~\cite{krizhevsky2012imagenet} & \multirow{8}{*}{Standard architecture and configurations} & 0.6808 & 0.8847 & 0.9203 & 58,417,305  \\
 VGG-11~\cite{simonyan2014very} &   & 0.6743 & 0.8814 & 0.9191 & 130,179,801 \\
 Inception V3~\cite{szegedy2016rethinking} &   & \red{0.7422} & \red{0.9189} & \red{0.9437} & 25,315,474  \\
 ResNet-18~\cite{he2016deep} &  & 0.7031 & 0.9030 & 0.9351 & 11,353,497  \\
 ResNet-34~\cite{he2016deep} &   & 0.7009 & 0.9010 & 0.9347 & 21,461,657  \\
 ResNet-152~\cite{he2016deep} &  & 0.6924 & 0.8973 & 0.9312 & 58,850,713 \\
 DenseNet-201~\cite{huang2017densely} &  & 0.7050 & 0.9013 & 0.9331 & 18,755,673  \\
 MobileNet V2~\cite{Sandler_2018_CVPR} &   & \blue{0.7310} & \blue{0.9161} & \blue{0.9429} & 2,665,817 \\
{SCNet~\cite{liu2020improving}} & & 0.7123 & 0.9026 & 0.9351 & 24,222,489 \\
%ResNet-18+BSConv-U~\cite{} & & & & & 1,623,199 \\
%\blue{ResNet-34+BSConv-U~\cite{haase2020rethinking}} & & 0.7192 & 0.9050 & 0.9346 & 2,786,335 \\
{ResNet-102+BSConv-U~\cite{haase2020rethinking}} & & 0.7172 & 0.9037 & 0.9334 & 7,029,791\\
%ResNet-18+BSConv-S~\cite{} & & & & & 1,000,767\\
%ResNet-34+BSConv-S~\cite{} & & & & & 1,612,031\\
%ResNet-102+BSConv-S~\cite{} & & & & & 3,864,831\\
 \hline
 \hline
 \textit{Vanilla} Transformer \cite{vaswani2017attention} & \small{$\hat{d}=256, L=4, I=8, Dropout=0.1$, Fully-connected graph} & 0.5249 & 0.7802 & 0.8486 & 14,029,401   \\
 MGT (Base) & \small{$\hat{d}=128, L=4, I=24, Dropout=0.1$, ${\bf A}^{1\textrm{-hop}},{\bf A}^{2\textrm{-hop}},{\bf A}^{\textrm{global}}$ graphs} & 0.7070 & 0.9030 & 0.9351 & 10,096,601  \\
 MGT (Large) & \small{$\hat{d}=256, L=4, I=24, Dropout=0.25$, ${\bf A}^{1\textrm{-hop}},{\bf A}^{2\textrm{-hop}},{\bf A}^{\textrm{global}}$ graphs} & \magenta{0.7280} & \magenta{0.9106} & \magenta{0.9387} & 39,984,729 \\
\hline
\end{tabular}
}
\end{center}
\end{table*}
%--------------------------------------------------------------------------

\subsection{Results}
\label{sec:results}

For fair comparison with RNN and CNN baselines at various parameter budgets, we implement two configurations of MGT: Base (10M parameters) and Large (40M parameters).
Additionally, we perform several ablation studies to evaluate the effectiveness of our multi-graph architecture and our sketch-specific input design.
Our main results are presented in Table \ref{table:comparison_with_cnn_rnn}. \\

\begin{table*}[!t]
\caption{Ablation study for multi-graph architecture of MGT. 
GT denotes single-graph variants of MGT. 
The $1^{st}$/$2^{nd}$~best results per column are indicated in \textcolor{red}{\textbf{red}}/\textcolor{blue}{blue}.
% \red{In the ``Recognition Accuracy'' column, the results in black and blue denotes the original results and the new results obtained by automatic early stop, respectively. The results in magenta are obtained by the new ReLU positions and new A global.}
$||$ denotes the logical union operation.
}

\label{table:ablation_study_on_multi_graph_design}
%\small
%\footnotesize
\scriptsize
%\tiny
\begin{center}
\resizebox{\textwidth}{!}{
\begin{tabular}{l || c | c | c | c | c | c | c | c | c | c  }
\hline
  \multirow{2}{*}{Network} & \multicolumn{6}{c|}{Configurations} & \multicolumn{3}{c|}{Recognition Accuracy} & \multirow{2}{*}{\tabincell{c}{Parameter \\ Amount}} \\
  \cline{2-10}
 & $G$ & Graph Structure & $I_{total}$ & $\hat{d}$ & $L$ & Dropout & acc.@1 & acc.@5 & acc.@10 &   \\
\hline
\hline
 GT \#1   & 1 & Fully-connected (\textit{vanilla}) & 8 & 256 & 4 & 0.10 & {0.5249} & {0.7802} & {0.8486} & 14,029,401   \\
 GT \#2   & 1 & Intra-stroke Fully-connected & 8 &  256 & 4 & 0.10 &  {0.6487} & {0.8697} & {0.9151} & 14,029,401  \\
 GT \#3   & 1 & Random (10\%) & 8 & 256 & 4 & 0.10 & {0.5271} & {0.7890} & {0.8589} & 14,029,401  \\
 GT \#4   & 1 & Random (20\%) & 8 & 256 & 4 & 0.10 & {0.5352} & {0.7945} & {0.8617} & 14,029,401  \\
 GT \#5   & 1 & Random (30\%) & 8 & 256 & 4 & 0.10 & {0.5322} & {0.7917} & {0.8588} & 14,029,401  \\
 \hline
 GT \#6   & 1 & ${\bf A}^{1\textrm{-hop}}$  & 8 & 256 & 4 & 0.10 & {0.7023} & {0.8974} & {0.9303} & 14,029,401   \\
 GT \#7   & 1 & ${\bf A}^{2\textrm{-hop}}$ & 8 & 256 & 4 & 0.10 & {0.7082} & {0.8999} & {0.9336} &  14,029,401  \\
 GT \#8   & 1 & ${\bf A}^{3\textrm{-hop}}$ & 8 & 256 & 4 & 0.10 & {0.7028} & {0.8991} & {0.9327} & 14,029,401   \\
 GT \#9   & 1 & ${\bf A}^{\textrm{global}}$ & 8 & 256 & 4 & 0.10 & {0.5488} &  {0.8009} & {0.8659} & 14,029,401  \\
 GT \#10   & 1 &  ${\bf A}^{1\textrm{-hop}}||{\bf A}^{2\textrm{-hop}}||{\bf A}^{\textrm{global}}$ & 8 & 256 & 4 & 0.10 & 0.7057 &  0.9021 & 0.9346 & 14,029,401  \\
%   GT \#9 \magenta{new A global}   & 1 & ${\bf A}^{\textrm{global}}$ & 8 & 256 & 4 & 0.10 & \magenta{0.5329} & \magenta{0.7886}  & \magenta{0.8545} & 14,029,401  \\
%   GT  & 1 & \magenta{${\bf A}^{1\textrm{-hop}},{\bf A}^{2\textrm{-hop}},{\bf A}^{\textrm{global}}$ sum} & 8 & 256 & 4 & 0.10 & \magenta{0.5201} & \magenta{0.7835}  & \magenta{0.8516} & 14,029,401  \\
 \hline
 MGT \#11   & 2 & ${\bf A}^{1\textrm{-hop}},{\bf A}^{2\textrm{-hop}}$ & 16 & 256  & 4 & 0.25 & {0.7149} & {0.9049 } & {0.9361} & 28,188,249   \\
%  MGT \#10 \magenta{new ReLU}   & 2 & ${\bf A}^{1\textrm{-hop}},{\bf A}^{2\textrm{-hop}}$ & 16 & 256  & 4 & 0.25 & \magenta{0.7093} & \magenta{0.9030} & \magenta{0.9348} & 28,188,249   \\
 MGT \#12   & 2 & ${\bf A}^{1\textrm{-hop}},{\bf A}^{\textrm{global}}$ & 16 & 256  & 4 & 0.25 & {0.7111} & {0.9041} & {0.9355} &  28,188,249   \\
%  MGT \#11 \magenta{new ReLU}   & 2 & ${\bf A}^{1\textrm{-hop}},{\bf A}^{\textrm{global}}$ & 16 & 256  & 4 & 0.25 & \magenta{0.7022} & \magenta{0.9005} & \magenta{0.9327} &  28,188,249   \\
%  MGT \#11 \magenta{new A global}   & 2 & ${\bf A}^{1\textrm{-hop}},{\bf A}^{\textrm{global}}$ & 16 & 256  & 4 & 0.25 & \magenta{0.7120} & \magenta{0.9035} & \magenta{0.9361} &  28,188,249   \\
 MGT \#13   & 2 & ${\bf A}^{2\textrm{-hop}},{\bf A}^{\textrm{global}}$ & 16 & 256  & 4  & 0.25 & \blue{0.7237} & \blue{0.9102} & \red{\textbf{0.9400}} & 28,188,249   \\
 \hline
 MGT \#14   & 3 & ${\bf A}^{1\textrm{-hop}},{\bf A}^{1\textrm{-hop}},{\bf A}^{1\textrm{-hop}}$ & 24 & 256  & 4 & 0.25 & {0.7077} & {0.9020} & {0.9340} & 39,984,729   \\
 MGT \#15   & 3 & ${\bf A}^{1\textrm{-hop}},{\bf A}^{2\textrm{-hop}},{\bf A}^{3\textrm{-hop}}$ & 24 & 256  & 4 & 0.25 & {0.7156} & {0.9066} & {0.9365} & 39,984,729 \\
 MGT \#16 & 3 & ${\bf A}^{1\textrm{-hop}}||{\bf A}^{2\textrm{-hop}}||{\bf A}^{\textrm{global}}$ & 24 & 256  & 4 & 0.25 &  0.7126 & 0.9051 & 0.9372  & 39,984,729   \\
 \hline
 \hline
MGT \#17 & 3 & ${\bf A}^{1\textrm{-hop}},{\bf A}^{2\textrm{-hop}},{\bf A}^{\textrm{global}}$ & 24 & 256  & 4 & 0.25 &  \red{\textbf{0.7280}} & \red{\textbf{0.9106}} & \blue{{0.9387}}  & 39,984,729   \\
% MGT \#15 & 3 & \magenta{${\bf A}^{1\textrm{-hop}},{\bf A}^{2\textrm{-hop}},{\bf A}^{\textrm{global}}$ sum} & 24 & 256  & 4 & 0.25 &  \magenta{0.5390} & \magenta{0.7955} & \magenta{0.8603}  & 39,984,729   \\
% MGT \#15 & 3 & ${\bf A}^{1\textrm{-hop}},{\bf A}^{2\textrm{-hop}},{\bf A}^{\textrm{global}}$ \magenta{new ReLU} & 24 & 256  & 4 & 0.25 &  \magenta{0.7216} & \magenta{0.9082} & \magenta{0.9380}  & 39,984,7 29   \\
% MGT \#15 & 3 & ${\bf A}^{1\textrm{-hop}},{\bf A}^{2\textrm{-hop}},{\bf A}^{\textrm{global}}$ \magenta{new ReLU, new A global} & 24 & 256  & 4 & 0.25 &  \magenta{0.7244} & \magenta{0.9109} & \magenta{0.9395}  & 39,984,729   \\
 \hline
\end{tabular}}
\end{center}
\end{table*}

%\vspace{3pt}
\subsubsection{Comparison with RNN Baselines}
%\noindent\textbf{Comparison with RNN Baselines }
% \paragraph{Comparison with RNN Baselines}
We trained RNNs at various parameter budgets, and present result for the best performing bi-directional LSTM and GRU models in Table \ref{table:comparison_with_cnn_rnn}:
(i) MGT outperforms both LSTM and GRU baselines by a significant margin (by 3\% acc.@1 for Base, 5\% for Large),
indicating that both geometry and temporal order of strokes are important for sketch representation learning.
(ii) Training larger RNNs is harder to converge, leading to degrading performance, \eg, GRUs outperform deeper LSTMs by 2\%.

These results are not surprising: RNNs are notoriously hard to train at scale \cite{pascanu2013difficulty}, 
while Transformer performance is known to improve with scale, even with billions of model parameters \cite{shoeybi2019megatron}. \\

\vspace{3pt}
\subsubsection{Comparison with CNN Baselines}
%\noindent\textbf{Comparison with CNN Baselines }
% \paragraph{Comparison with CNN Baselines}
Table \ref{table:comparison_with_cnn_rnn} also presents performance of several state-of-the-art CNN architectures for computer vision:
(i) \black{Inception V3 \cite{szegedy2016rethinking} and MobileNet V2~\cite{Sandler_2018_CVPR} are the best performing CNN architectures. Our MGT Base has competitive or better recognition accuracy than all other baselines: AlexNet \cite{krizhevsky2012imagenet}, VGG-11 \cite{simonyan2014very}, ResNet models \cite{he2016deep}, and DenseNet-201 \cite{huang2017densely}.}
(ii) \black{MGT Large has small performance gap to Inception V3 and MobileNet V2 (\ie, 72.80\% acc.@1 vs. 74.22\%, 72.80\% acc.@1 vs. 73.10\%) and outperforms all other CNN architectures by almost 2\%.}
(iii) Somewhat counter-intuitively, shallow networks (Inception V3, MobileNet V2) outperform deeper networks (ResNet-152, Densenet-201) by almost 2\%.
This result highlights that CNNs designed for images with dense colors and textures are un-suitable for sparse sketches.

Note that MobileNet V2 is specifically designed for fast inference on mobile phones and is not directly comparable in terms of model parameters.
\\

\begin{table*}[!tbp]
\caption{Test set performance of Graph Transformer vs. other GNN variants.
The $1^{st}$/$2^{nd}$~best results per column are indicated in \textcolor{red}{\textbf{red}}/\textcolor{blue}{blue}.
%``fully'' denotes fully-connected.
%``Gra. Stru.'' denotes graph structure.
}

\label{table:comparison_with_gnn}
%\small
%\footnotesize
%\scriptsize
%\tiny
\begin{center}
\resizebox{0.85\textwidth}{!}{
\begin{tabular}{l || c | c | c | c | r }
\hline
  \multirow{2}{*}{Network} & \multirow{2}{*}{Graph Structure} & \multicolumn{3}{c|}{Recognition Accuracy} & \multirow{2}{*}{\tabincell{c}{Parameter \\ Amount}}  \\
  \cline{3-5}
 & & acc.@1 & acc.@5 & acc.@10 &   \\
\hline
\hline
\multirow{2}{*}{Graph
  Convolutional Networks (GCN) \cite{kipf2017semi}} & fully-connected & 0.4098 & 0.7384 & 0.8213 & \multirow{2}{*}{6,948,441} \\ 
 & ${\bf A}^{1\textrm{-hop}}$ & 0.6800 & 0.8869 & 0.9224 &  \\
\hline
\multirow{2}{*}{Graph Attention Networks (GAT) \cite{velickovic2018graph}} & fully-connected & 0.4098 & 0.6960 & 0.7897 & \multirow{2}{*}{11,660,889} \\
 & ${\bf A}^{1\textrm{-hop}}$ & \blue{0.6977} & \blue{0.8952} & \blue{0.9298} &  \\
\hline
\hline
\multirow{2}{*}{Graph Transformer (GT)} & fully-connected & 0.5242 & 0.7796 & 0.8465 & \multirow{2}{*}{14,029,401} \\
 & ${\bf A}^{1\textrm{-hop}}$ & \red{\textbf{0.7057}} & \red{\textbf{0.8992}} & \red{\textbf{0.9311}} &  \\
\hline
position-wise feed-forward & None & 0.5296 & 0.7901 & 0.8576 & 4,586,073 \\
% \hline
% GT (Ours) & 5-NN & 0.5934 & 0.8366 & 0.8904 & 14,029,401 \\
% GT (Ours) & 10-NN & 0.5877 & 0.8340 & 0.8878 & 14,029,401 \\
% GT (Ours) & 25-NN & 0.5649 & 0.8102 & 0.8703 & 14,029,401 \\
\hline
\end{tabular}
}
\end{center}
\end{table*}

\begin{table*}[!tb]
\caption{Ablation study for multi-modal input for MGT (Large). Notations: ``+'' and ``$\mathcal{C}(\cdots)$'' denote ``sum'' and ``concatenate'', respectively.
The $1^{st}$/$2^{nd}$~best results per column are indicated in \textcolor{red}{\textbf{red}}/\textcolor{blue}{blue}.
}
\label{table:comparison_on_inputs}
%\small
%\footnotesize
%\scriptsize
\tiny
\begin{center}
\resizebox{0.7\textwidth}{!}{
\begin{tabular}{l || c | c | c  }
\hline
  \multirow{2}{*}{Input Permutation} & \multicolumn{3}{c}{Recognition Accuracy} \\
  \cline{2-4}
 & ~acc.@1~ & ~acc.@5~ & ~acc.@10~  \\
\hline
\hline
 coordinate & 0.6512 & 0.8735 & 0.9162   \\
 coordinate + flag bit &  0.6568 & 0.8762 & 0.9176 \\
 coordinate + flag bit + position encoding & 0.6600 & 0.8766 &  0.9182   \\
 $\mathcal{C}$(coordinate, flag bit) & 0.7017 & 0.8996 & 0.9321  \\
 $\mathcal{C}$(coordinate, flag bit, position encoding) & \red{\textbf{0.7280}} & \red{\textbf{0.9106}} & \red{\textbf{0.9387}}    \\
 \hline
 4D Input & 0.6559 & 0.8758 & 0.9175 \\
 4D Input + position encoding & 0.6606 & 0.8781 & 0.9190   \\
 $\mathcal{C}$(4D Input, position encoding) & \blue{0.7117} & \blue{0.9048} & \blue{0.9366}  \\
%  \hline
%  \hline
%  coo.  & BASE & 0.6207 & 0.8533 & 0.9030   \\
%  coo. + flag  & BASE &  0.6237 &  0.8561 & 0.9062 \\
%  coo. + flag + pos.  & BASE & 0.6273 &  0.8582 & 0.9061   \\
%  $\mathcal{C}$(coo., flag) & BASE & 0.6857 & 0.8911 & 0.9272  \\
%  $\mathcal{C}$(coo., flag, pos.) & BASE & 0.7070 & 0.9030 & 0.9351 \\
%  \hline
%  4D input  &  BASE  &  0.6251 & 0.8575 &  0.9063  \\
%  4D input + pos.   & BASE & 0.6303 & 0.8594 & 0.9081   \\
%  $\mathcal{C}$(4D input, pos.)  & BASE & 0.6926 & 0.8962 & 0.9308  \\
 \hline
 
\end{tabular}
}
\end{center}
\end{table*}

%----------------------------------------------------------------------------------------

%\begin{table*}[!tb]
%\caption{Ablation study for multi-modal input for MGT (Large). Notations: ``+'' and ``$\mathcal{C}(\cdots)$'' denote ``sum'' and ``concatenate'', respectively; ``coo.'', ``flag'', and ``pos.'' represent ``coordinate'', ``flag bit'', and ``position encoding'', respectively.
%The $1^{st}$/$2^{nd}$~best results per column are indicated in \textcolor{red}{\textbf{red}}/\textcolor{blue}{blue}.
%}
%\label{table:comparison_on_inputs}
%%\small
%%\footnotesize
%%\scriptsize
%%\tiny
%\begin{center}
%\resizebox{0.7\textwidth}{!}{
%\begin{tabular}{l || c | c | c  }
%\hline
%  \multirow{2}{*}{Input Permutation} & \multicolumn{3}{c}{Recognition Accuracy} \\
%  \cline{2-4}
% & ~acc.@1~ & ~acc.@5~ & ~acc.@10~  \\
%\hline
%\hline
% coordinate & 0.6512 & 0.8735 & 0.9162   \\
% coordinate + flag bit &  0.6568 & 0.8762 & 0.9176 \\
% coordinate + flag bit + position encoding & 0.6600 & 0.8766 &  0.9182   \\
% $\mathcal{C}$(coordinate, flag bit) & 0.7017 & 0.8996 & 0.9321  \\
% $\mathcal{C}$(coordinate, flag bit, position encoding) & \red{\textbf{0.7280}} & \red{\textbf{0.9106}} & \red{\textbf{0.9387}}    \\
% \hline
% 4D Input & 0.6559 & 0.8758 & 0.9175 \\
% 4D Input + position encoding & 0.6606 & 0.8781 & 0.9190   \\
% $\mathcal{C}$(4D Input, position encoding) & \blue{0.7117} & \blue{0.9048} & \blue{0.9366}  \\
% \hline
% 
%\end{tabular}
%}
%\end{center}
%\end{table*}

%----------------------------------------------------------------------------------------

% \vspace{3pt}
% \newpage
\subsubsection{Ablations for Multi-Graph Architecture}
%\noindent\textbf{Ablations for Multi-Graph Architecture }
% \paragraph{Ablations for Multi-Graph Architecture}
We design several ablation studies to evaluate our sketch-specific multi-graph architecture in Table \ref{table:ablation_study_on_multi_graph_design}:
(i) We evaluate Graph Transformers trained on fully-connected graphs, \textit{i.e.} \textit{vanilla} Transformers (GT \#1), fully-connected graphs within strokes (GT \#2), as well as random graphs with 10\%, 20\% and 30\% connectivity (GT \#3, \#4, and \#5 respectively).
We compare their performance with Graph Transformers trained on sketch-specific graphs ${\bf A}^{1\textrm{-hop}}$ (GT \#6), ${\bf A}^{2\textrm{-hop}}$ (GT \#7), ${\bf A}^{3\textrm{-hop}}$ (GT \#8), and ${\bf A}^{\textrm{global}}$ (GT \#9).
We find that \textit{vanilla} Transformers on fully-connected (52.49\% acc.@1) and random graphs (52.71\%, 53.52\%, 53.22\%) perform poorly compared to sketch-specific graph structures determined by domain expertise, such as fully-connected stroke graphs (64.87\%) and ${\bf A}^{1\textrm{-hop}}$ (70.23\%).
The superior performance of $K$-hop graphs suggests that Transformers benefit from sparse graphs representing local sketch geometry.
\black{
We also evaluate a combined sketch-specific graph structure, \ie, ${\bf A}^{1\textrm{-hop}}||{\bf A}^{2\textrm{-hop}}||{\bf A}^{\textrm{global}}$ (GT \#10),  where the graph connectivity is the logical union set of ${\bf A}^{1\textrm{-hop}}$, ${\bf A}^{2\textrm{-hop}}$, and ${\bf A}^{\textrm{global}}$.
However, this structure fails to gain performance improvement over ${\bf A}^{1\textrm{-hop}}$, ${\bf A}^{2\textrm{-hop}}$, and ${\bf A}^{\textrm{global}}$, despite involving more domain knowledge.
}

(ii) We experiment with various permutations of graphs for multi-graph models (MGT \#11-\#17).
We find that using a 3-graph architecture (MGT \#17) combining local sketch geometry (${\bf A}^{1\textrm{-hop}}, {\bf A}^{2\textrm{-hop}}$) and global temporal relationships (${\bf A}^{\textrm{global}}$) significantly boosts performance over 2-graph and 1-graph models (72.80\% vs. 72.37\% for 2-graph and 70.82\% for 1-graph).
This result is interesting because using global graphs independently (GT \#9) leads to comparatively poor performance (54.88\%).
Additionally, we found that using diverse graphs (MGT \#15, \#17) is better than using the same graph (MGT \#14).
Comparing MGT \#14 and MGT \#6 further shows that performance gains are due to the multi-graph architecture as opposed to more model parameters.

(iii) We also repeatedly input the adjacency matrix of GT \#10 (\ie, ${\bf A}^{1\textrm{-hop}}||{\bf A}^{2\textrm{-hop}}||{\bf A}^{\textrm{global}}$) three times as the multiple graph structures to train our MGT (see MGT \#16 in Table~\ref{table:ablation_study_on_multi_graph_design}). Compared with MGT \#17, there is a clear performance gap (71.26\% vs. 72.80\%).
This further validates our idea of learning sketch representations through multiple separate graphs.\\

% \vspace{3pt}
\subsubsection{Comparison with GNN Baselines}
%\noindent\textbf{Comparison with GNN Baselines }
% \paragraph{Comparison with GNN Baselines}
In Table \ref{table:comparison_with_gnn}, we present performance of our Graph Transformer model compared to Graph
  Convolutional Networks (GCN) \cite{kipf2017semi} and Graph Attention Networks~(GAT)~\cite{velickovic2018graph}, two popular GNN variants:
(i) We find that all models perform similarly on fully-connected graphs.
Using 1\textrm{-hop} graphs results in significant gains for all models, with Transformer performing the best.
(ii) Interestingly, both GNNs on fully-connected graphs are outperformed by a simple position-wise embedding method without any graph structure: 
each node undergoes 4 feed-forward (FF) layers followed by summation and the MLP classifier.
These results further highlights the importance of sketch-specific graph structures for the success of Transformers.
Our final models use the Transformer layer, which implicitly includes the FF sub-layer \eqref{equ:mgt}. \\

% \begin{table}[!tbp]
% \caption{Performance of best baseline models vs. MGT on $1235$ truncated sketches from the test set. 
% The $1^{st}$/$2^{nd}$~best results per column are indicated in \textcolor{red}{\textbf{red}}/\textcolor{blue}{blue}.
% }
% \label{table:comparison_on_1235_truncated}
% %\small
% %\footnotesize
% %\scriptsize
% %\tiny
% \begin{center}
% \resizebox{\columnwidth}{!}{
% \begin{tabular}{l || c | c | c  }
% \hline
%   \multirow{2}{*}{Network} &  \multicolumn{3}{c}{Recognition Accuracy} \\
%   \cline{2-4}
%   & acc.@1 & acc.@5 & acc.@10  \\
% \hline
% \hline
%  Bi-directional GRU & 0.4481 & 0.7143 & 0.7684  \\
%  Inception V3~\cite{szegedy2016rethinking}      & \red{\textbf{0.5336}} & \red{\textbf{0.7603}} & \red{\textbf{0.7976}}   \\
%  \hline
%  \hline
%  MGT (Large) & \blue{0.4591} & \blue{0.7182} & \blue{0.7879}  \\
% \hline
% \end{tabular}
% }
% \end{center}
% \end{table}

% \begin{figure}[!t]
% \begin{center}
% %\fbox{\rule{0pt}{2in} \rule{.9\linewidth}{0pt}}
% \includegraphics[width=0.9\columnwidth]{./figs/RNN_GT_comparison_on_different_stroke_ranges_v20191220.pdf}
% \end{center}
%   \caption{Test set performance of best baseline models vs. MGT, visualized by number of key points per sketch. Truncated samples are excluded.}
% \label{fig:RNN_GT_comparison_on_different_stroke_ranges}
% % \vspace{-0.3cm}
% \end{figure}

% \vspace{3pt}
\subsubsection{Ablations for Multi-Modal Input}
%\noindent\textbf{Ablations for Multi-Modal Input }
% \paragraph{Ablations for Multi-Modal Input}
In Table \ref{table:comparison_on_inputs}, we experiment with various permutations of our sketch-specific multi-modal input design.
We aggregate information from spatial (coordinates), semantic (flag bits), and temporal (position encodings) modalities via summation (as in Transformers for NLP) or concatenation:
(i) Effectively using all modalities is important for performance (\eg, ``$\mathcal{C}$(coordinate, flag bit, position encoding)'' outperforms ``coordinate'' and ``$\mathcal{C}$(coordinate, flag bit)'': 72.80\% acc.@1 vs. 65.12\%, 70.17\%). 
(ii) Concatenation works better than 4D input as well as summation (\eg, ``$\mathcal{C}$(coordinate, flag bit, position encoding)'' outperforms ``$\mathcal{C}$(4D Input, position encoding)'' and ``coordinate + flag bit + position encoding'': 72.80\% vs. 71.17\%, 66.06\%). \\

% \blue{
% \vspace{3pt}
% \noindent\textbf{Performance on Truncated Sketches }
% % \paragraph{Performance on Truncated Sketches}
% CNNs can be seen as an upper bound for RNNs and MGT due to truncation of sketch sequence during training.
% In Table \ref{table:comparison_on_1235_truncated}, we evaluate the best performing RNN, CNN and MGT models on $1,235$ truncated sketches from our test set, and find that sketch truncation skews results in favour of CNNs over RNNs/MGT, as CNNs operate on complete images of truncated sketches.
% \\
% Furthermore, visualizing model performance based on key points per sketch (Figure \ref{fig:RNN_GT_comparison_on_different_stroke_ranges})
% clearly shows that MGT matches/outperforms CNN baselines on non-truncated sketches.
% In future work, we shall explore more powerful graph aggregation strategies, 
% % such as Memory Networks \cite{sukhbaatar2015end} or attention \cite{knyazev2019understanding}, 
% for improving MGT on complex and truncated sketches.
% }

%--------------------------------
% \vspace{3pt}

%---------------------------------------------------------------------------
\begin{figure*}[!t]
    \centering
    \subfigure[${\bf A}^{1\textrm{-hop}}$]{
        \includegraphics[width=0.76\textwidth]{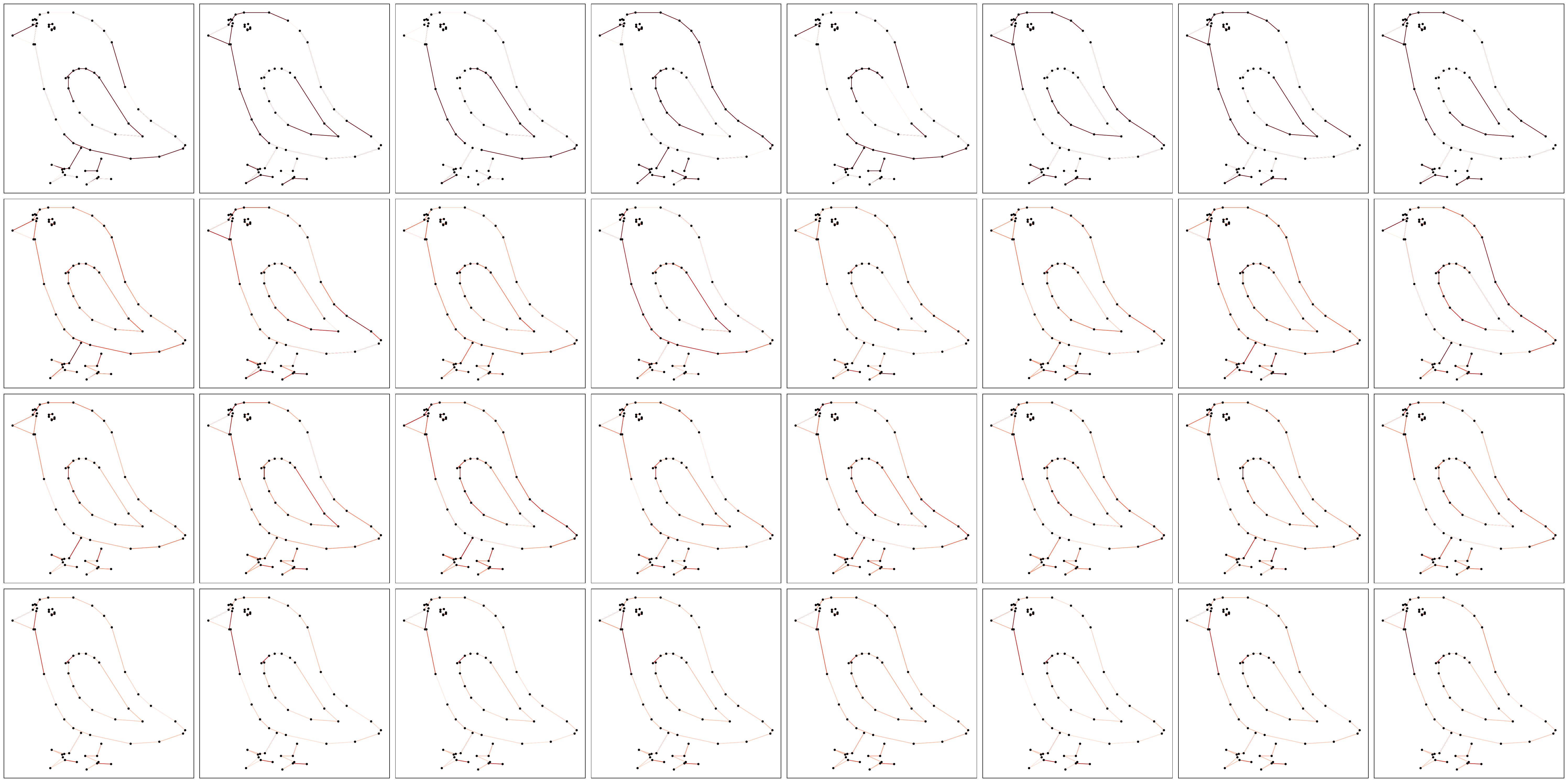}
    }
    \subfigure[${\bf A}^{2\textrm{-hop}}$]{
        \includegraphics[width=0.76\textwidth]{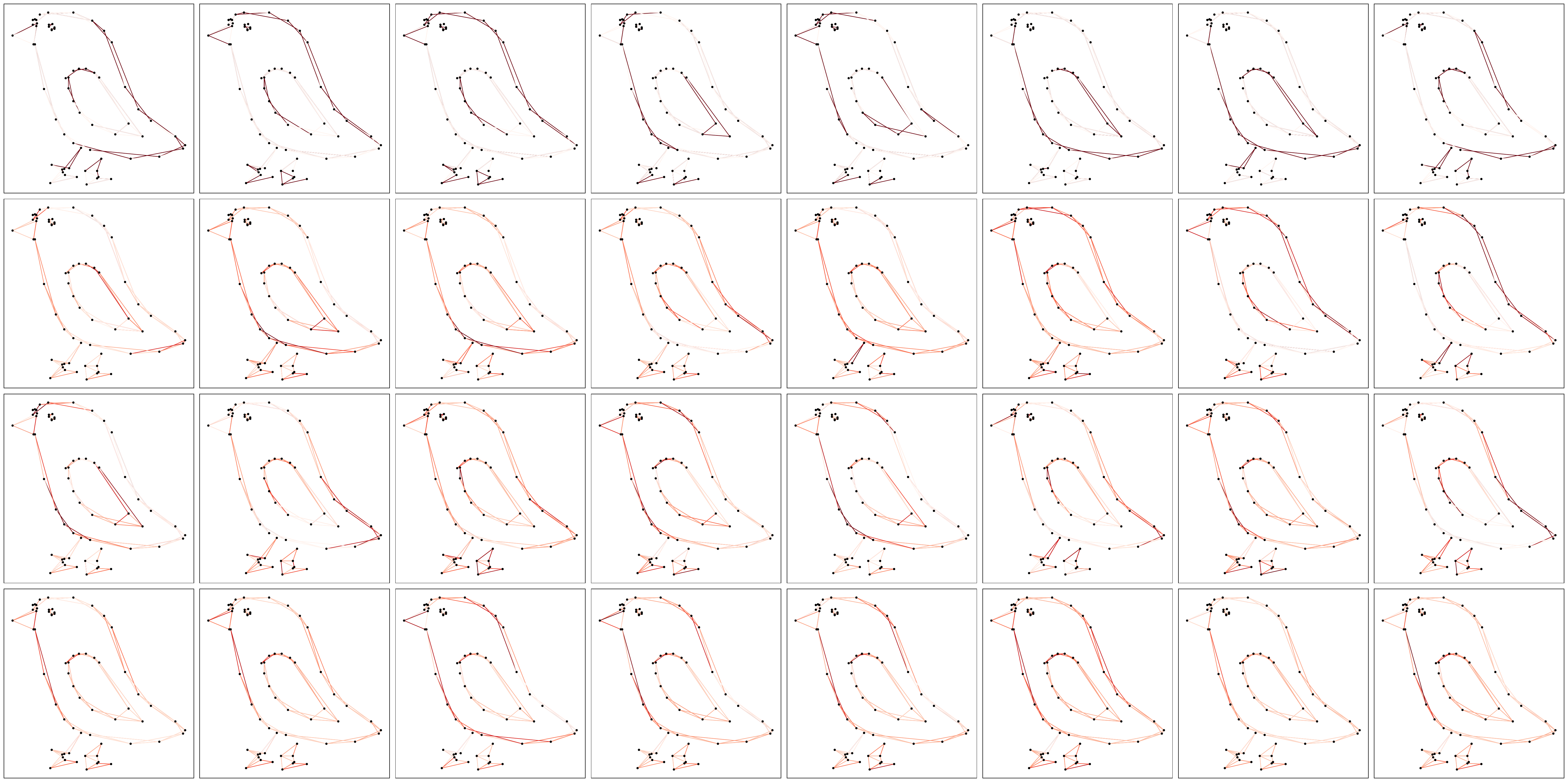}
    }
    \subfigure[${\bf A}^{global}$]{
        \includegraphics[width=0.76\textwidth]{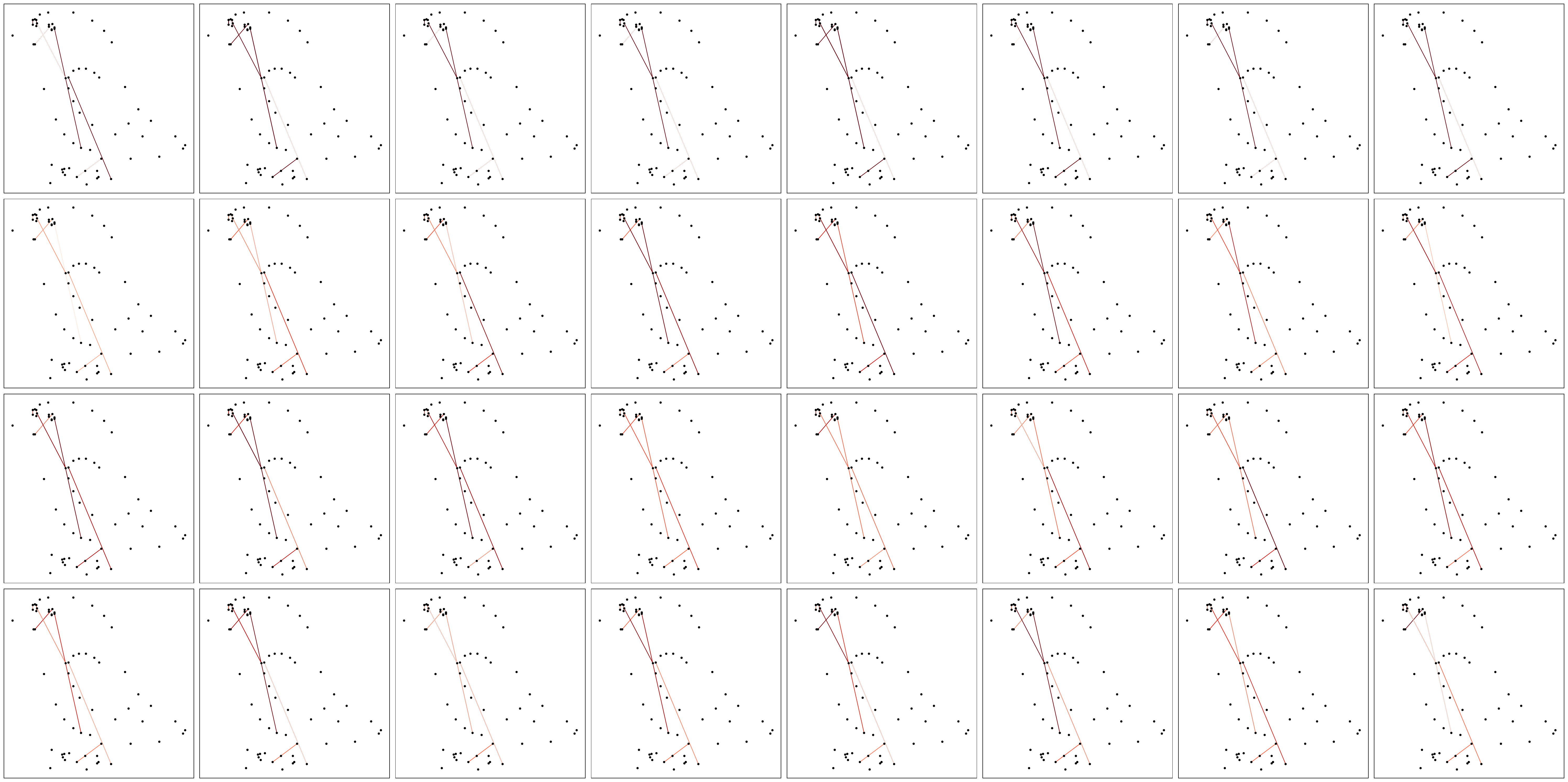}
    }
    \caption{
    Attention heads at each layer of MGT for a test set sample labelled \textit{bird}. Each layer has $I=8$ attention heads per graph in total. Darker reds indicate higher attention values. Best viewed in color.
    }
    \label{fig:sup-viz-bird}
\end{figure*}

%---------------------------------------------------------------------------

\begin{figure*}[!t]
    \centering
    \subfigure[${\bf A}^{1\textrm{-hop}}$]{
        \includegraphics[width=0.76\textwidth]{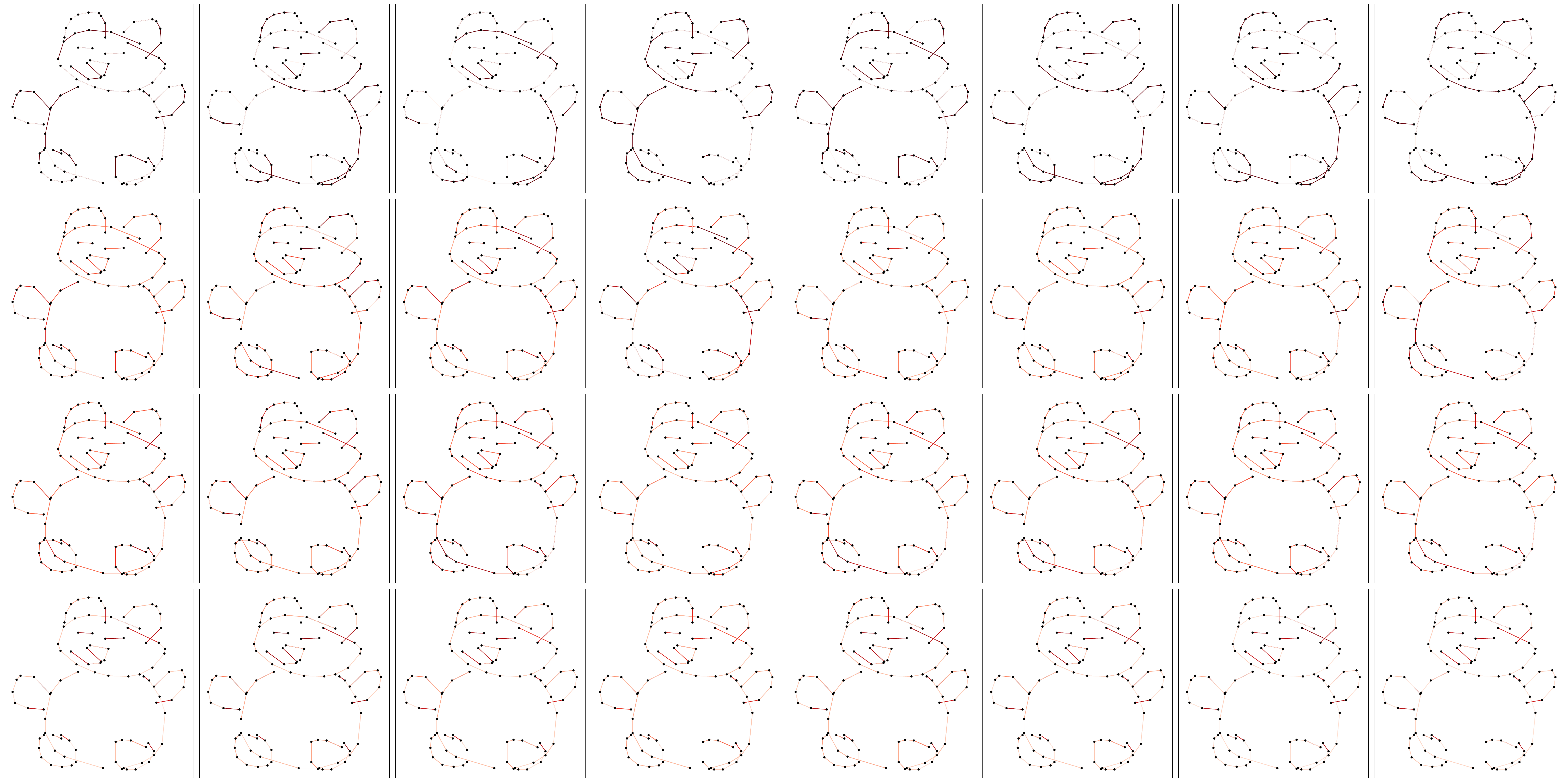}
    }
    \subfigure[${\bf A}^{2\textrm{-hop}}$]{
        \includegraphics[width=0.76\textwidth]{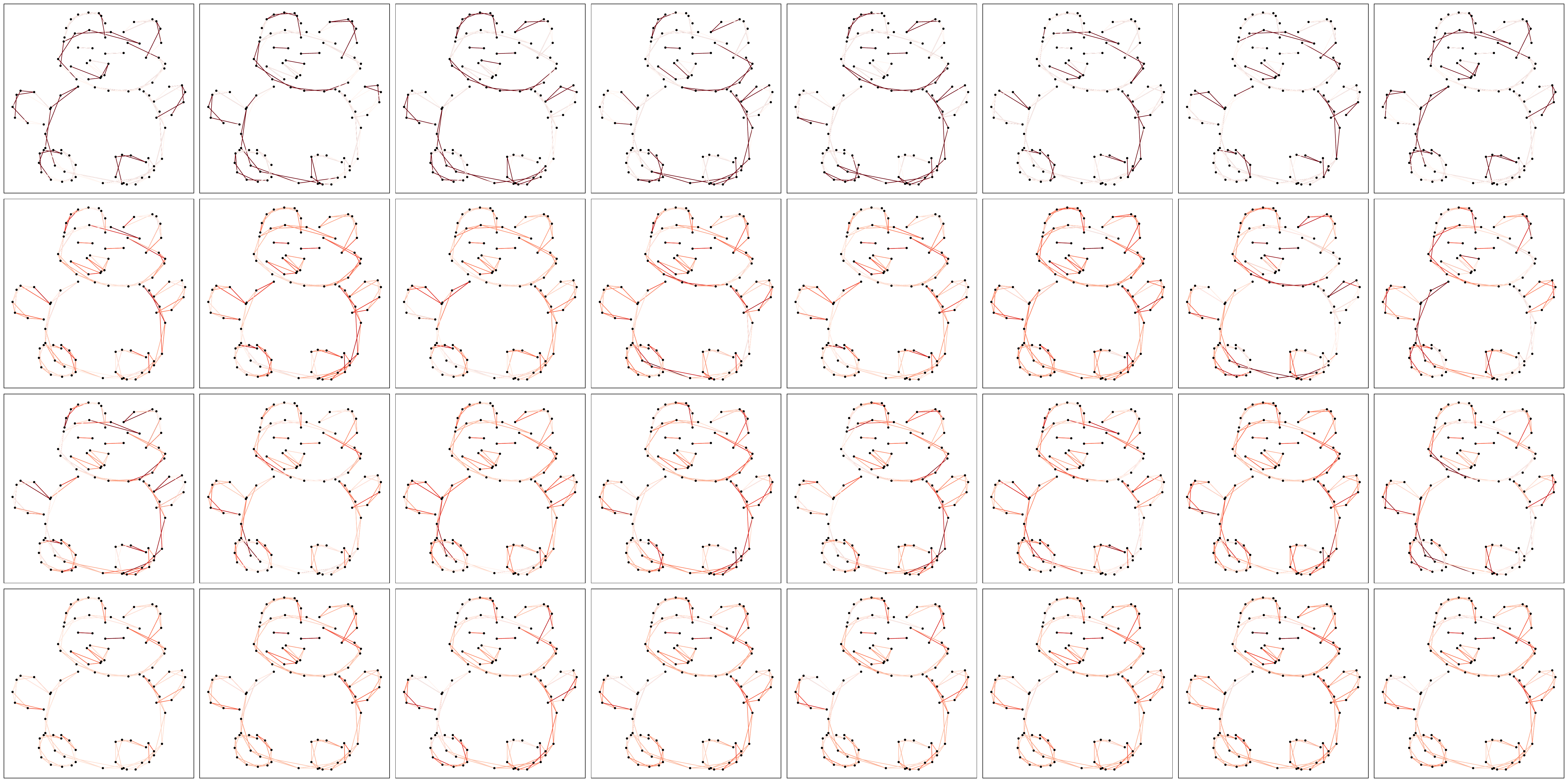}
    }
    \subfigure[${\bf A}^{global}$]{
        \includegraphics[width=0.76\textwidth]{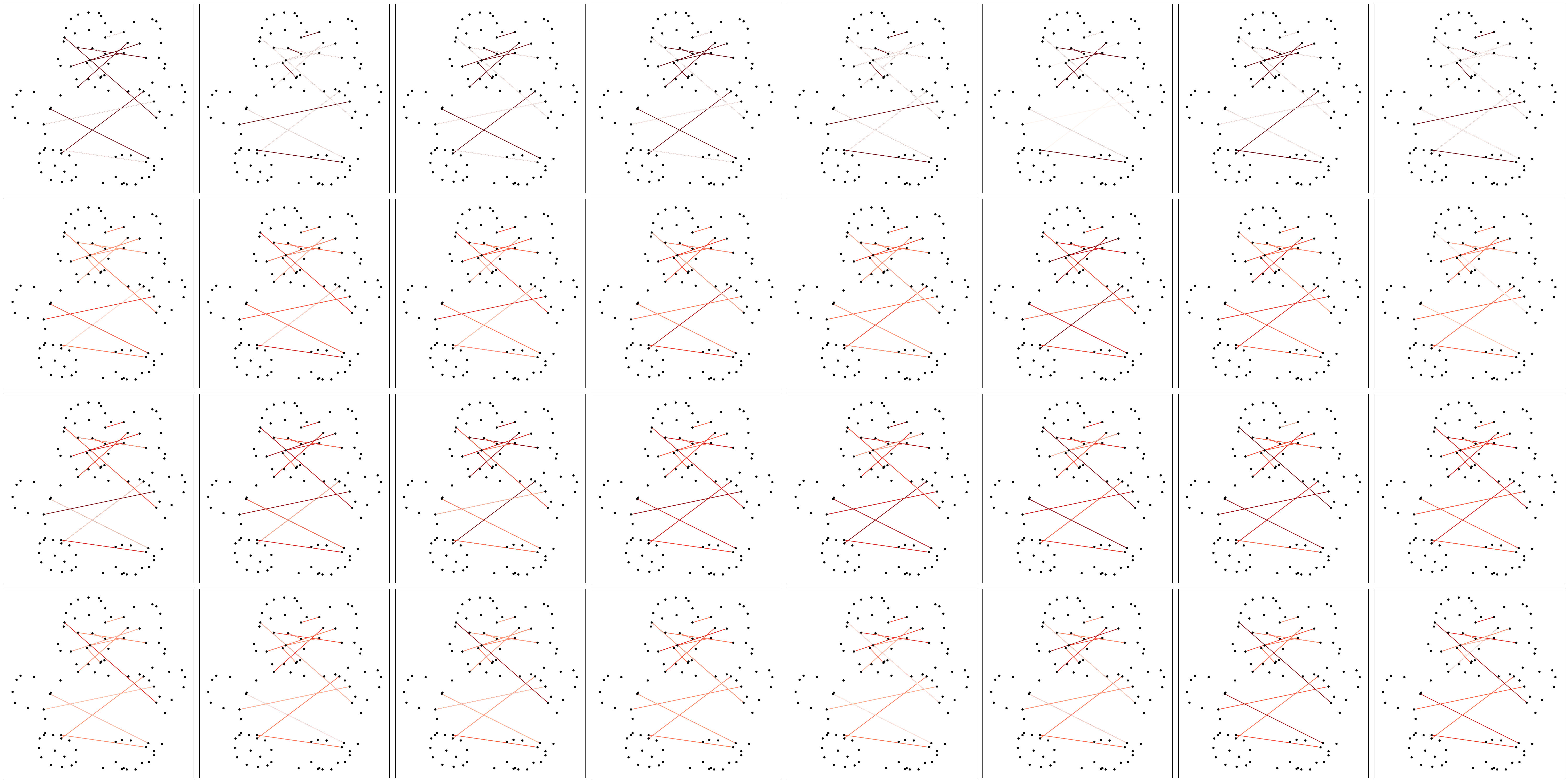}
    }
    \caption{
    Attention heads at each layer of MGT for a test set sample labelled \textit{teddy}. Each layer has $I=8$ attention heads per graph in total. Darker reds indicate higher attention values. Best viewed in color.
    }
    \label{fig:sup-viz-teddy}
\end{figure*}

%----------------------------------------------------------------------------

\subsubsection{Qualitative Results}
%\noindent\textbf{Qualitative Results }
% \paragraph{Qualitative Results}
%In Figure \ref{fig:visualization_fig2}, we visualize attention heads at each layer of MGT for a sample from the test set (labelled \textit{`alarm clock'}).
\black{
In Figure~\ref{fig:sup-viz-bird} and Figure~\ref{fig:sup-viz-teddy}, we visualize attention heads at each layer of MGT for various test set samples.
Each sub-figure contains attention heads for each of the three graphs (${\bf A}^{1\textrm{-hop}}$, ${\bf A}^{2\textrm{-hop}}$, ${\bf A}^{global}$),
and each of the rows \#1-\#4 in each sub-figure correspond to layers \#1-\#4.
Darker reds indicate higher attention values. All figures are best viewed in color.
}

Attention heads in the initial layers attend very strongly to certain neighbors and very weakly to others,
\ie, the model builds local patterns for sketch sub-components (strokes) through message passing along their contours.
In penultimate layers, the intensity of neighborhood attention is significantly lower and evenly distributed, 
indicating that the model is aggregating information from various strokes at each node.

Additionally, we believe $A^{\textrm{global}}$ graphs are critical for message passing between strokes,
enabling the model to understand their relative arrangement. 
\black{
For example, in Figure~\ref{fig:sup-viz-bird}, both the head and feet of the bird are attached to the bottom of its body.
In Figure~\ref{fig:sup-viz-teddy}, the feet of the teddy bear are associated.  
}\\
% More visualizations are provided in the Supplementary Material.

%----------------------------------------------------------------

\begin{table}[!t]
\caption{
Evaluation timing comparison for Multi-Graph Transformer and CNNs, averaged over 3 runs.
Inference time is the total wall clock time for performing inference over 34,500 sketches from the test set.
%Performance comparison on the relation extraction task on SemEval-2010 task 8.
%The $1^{st}$/$2^{nd}$~best results per column are indicated in \textcolor{red}{\textbf{red}}/\textcolor{blue}{blue}.
}
\label{table:comparison_on_time_cost}
%\small
%\footnotesize
%\scriptsize
\tiny
\begin{center}
\resizebox{0.8\columnwidth}{!}{
\begin{tabular}{l || c | r }
\hline
  Network & Inference Time & \tabincell{c}{Parameter \\ Amount} \\
\hline
\hline
Inception V3 & 2.682$\pm$0.154s & 25,315,474 \\
MobileNet V2 & 1.423$\pm$0.070s & 2,665,817 \\
\hline
{MGT (Base)} & 0.849$\pm$0.044s & 10,096,601 \\
{MGT (Large)} &  0.843$\pm$0.048s & 39,984,729 \\
%\hline
%\multirow{2}{*}{MGT (serial implementation)} & 1.891$\pm$0.124s & 10,096,601 \\
% &  2.136$\pm$0.034s & 39,984,729 \\
 \hline
 
\end{tabular}
}
\end{center}
\end{table}

%----------------------------------------------------------------
%----------------------------------------------------------------

\begin{table}[!t]
\caption{Performance comparison on the relation extraction task on SemEval-2010 task 8.
The $1^{st}$/$2^{nd}$~best results per column are indicated in \textcolor{red}{\textbf{red}}/\textcolor{blue}{blue}.
}
\label{table:comparison_on_relation_extract}
%\small
%\footnotesize
%\scriptsize
%\tiny
\begin{center}
\resizebox{\columnwidth}{!}{
\begin{tabular}{l || c | c }
\hline
  Network & Graph Structure & \tabincell{c}{Prediction\\ Accuracy} \\
\hline
\hline
textual CNN~\cite{guo2019single}  & - & 0.8660 \\
Transformer~\cite{wang2019extracting} & fully-connected & \textcolor{blue}{0.8900} \\
Transformer & fully- and partially-connected & \textcolor{red}{\textbf{0.8945}} \\
 \hline
 
\end{tabular}
}
\end{center}
\end{table}

%----------------------------------------------------------------

\subsubsection{Comparison for Time Cost}
\label{sec:timecost}
%\noindent\textbf{Comparison for Time Cost}
To demonstrate the speed advantage of our model, we also compare our MGT with the strongest CNN baselines, \ie, Inception V3, MobileNet V2.
We report the total time taken by models to perform inference over 34,500 sketches from the test set in Table~\ref{table:comparison_on_time_cost}.
All models were implemented in PyTorch and run on an Intel Xeon CPU E5-2690 v4 server using a single Nvidia 1080Ti GPU, with batch size of 256 and 16 workers per run.

In Table~\ref{table:comparison_on_time_cost}, we observe:
%(i) The serial implementation of our MGT infers faster than Inception V3 and slower than MobileNet V2.
%(ii) The parallel implementation of our MGT infers obviously faster than both Inception V3 and MobileNet V2. 
{our MGT infers obviously faster than both Inception V3 and MobileNet V2}. 

Note that, unlike CNNs and RNNs, GNNs and GTs for sparse graph data formats are not natively supported by PyTorch.
We believe that our MGT can infer faster in our future work, if we further speed up MGT via using the tailor-made GNN libraries such as Deep Graph Library~\footnote{\url{https://www.dgl.ai/}} and PyTorch Geometric~\footnote{\url{https://github.com/rusty1s/pytorch_geometric}}. \\

\subsubsection{Evaluation on Relation Extraction}
\begin{figure}[!t]
    \centering
    \includegraphics[width=0.9\columnwidth]{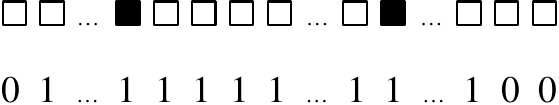}
    \caption{{A partially-connected graph illustration for a  long sentence. The entities and the other words are denoted by the solid and hollow squares, respectively. The corresponding attention mask is also provided.}}
    \label{fig:re}
\end{figure}

%\noindent\textbf{Evaluation on Relation Extraction}
The main idea of our multi-graph transformer is injecting domain knowledge into Transformers through domain-specific graphs.
We also try to evaluate this idea in other modalities  beyond sketch, \eg, NLP tasks.
Thus, 
we transfer our multi-graph transformer idea to conduct Relation Extraction (RE) on a RE benchmark (SemEval-2010 task 8~\cite{hendrickx2010semeval}), outperforming the state-of-the-art CNNs by a clear margin (as reported in Table~\ref{table:comparison_on_relation_extract}, prediction accuracy: ours (89.45\%) vs. CNN~\cite{guo2019single} (86.60\%)). 
In particular,
when we use Bidirectional Encoder Representations from Transformers (BERT)~\cite{devlin2018bert} as Transformer-based encoder, each sentence  will be encoded as a fully-connected graph.
Given a long sentence, if its two entities are overly far from each other, the RE model should give more attention on the words between and around the two entities.
Therefore, we also use multiple graph structures to inject this domain-knowledge into the Transformer-based encoder:
(i) Fully-connected: In the early and middle stages of training, we allow the transformer to encode each sentence as a fully-connected graph, regardless of its length. 
(ii) Partially-connected: In the last few epochs, if in a given long sentence its two entities are far apart, we force the transformer to encode the long sentence as a partially-connected graph, which consists of the words between and around the two entities. {We use a binary attention mask to encode this. See Figure~\ref{fig:re} for an illustration.} 

This experimental phenomenon further encourages us to explore the applications of using domain-specific graph structures to inject the domain-knowledge into Transformers in other modalities and tasks.

%\subsubsection{Advantages of Multi-Graph Transformer to CNN}

%------------------------------------------------------------------------

%---------------------------------------------------------
\section{Conclusion}
\label{sec:conclusion}

This paper introduces a novel representation of free-hand sketches as multiple sparsely connected graphs.
We design a Multi-Graph Transformer (MGT) for capturing both geometric structure and temporal information from sketch graphs.
\black{
The intrinsic traits of the MGT architecture include:}
(i) using graphs as universal representations of sketch geometry, as well as temporal and semantic information,
(ii) injecting domain knowledge into Transformers through sketch-specific graphs,
and
(iii) making full use of multiple intra-stroke and extra-stroke graphs.
% Based on our experimental results, we can draw several conclusions from the views of both sketch and graph: (i) Sketches can be well represented as graphes. (ii) The domain-expertise based graph structures are important to graph-based modeling. (iii) Representing a sketch graph as multiple graphs helps graph neural network to learn more patterns from it.

\black{For sketch community: We, for the first time, propose to model sketches as sparsely connected graphs. For GNN community: We design a novel graph transformer model that considers prior/domain knowledge via multiple graph structures. This multi-graph modeling idea works well for both sketch and other modalities.}

We hope MGT can serve as a foundation for future work in sketch applications and network architectures,
% steering 
motivating
the community towards sketch representation learning using graphs.
Additionally, for the graph neural network (GNN) community, we hope that MGT helps free-hand sketch become a new test-bed for GNNs.

%-----------------------------------------------------------------------
\section{Future Work}
\label{sec:future_work}
\black{
Our future work will focus on applying our MGT as a general neural representation in various sketch tasks, \eg, sketch generation. More code and results will be updated continuously  on our project page~\footnote{\url{https://github.com/PengBoXiangShang/multigraph_transformer}}.
}

%\section*{Acknowledgements}

%Xavier Bresson is supported in part by NRF Fellowship NRFF2017-10.

{\small
\bibliographystyle{IEEEtran}
\bibliography{egbib_modified}

\begin{thebibliography}{10}
\providecommand{\url}[1]{#1}
\csname url@samestyle\endcsname
\providecommand{\newblock}{\relax}
\providecommand{\bibinfo}[2]{#2}
\providecommand{\BIBentrySTDinterwordspacing}{\spaceskip=0pt\relax}
\providecommand{\BIBentryALTinterwordstretchfactor}{4}
\providecommand{\BIBentryALTinterwordspacing}{\spaceskip=\fontdimen2\font plus
\BIBentryALTinterwordstretchfactor\fontdimen3\font minus
  \fontdimen4\font\relax}
\providecommand{\BIBforeignlanguage}[2]{{%
\expandafter\ifx\csname l@#1\endcsname\relax
\typeout{** WARNING: IEEEtran.bst: No hyphenation pattern has been}%
\typeout{** loaded for the language `#1'. Using the pattern for}%
\typeout{** the default language instead.}%
\else
\language=\csname l@#1\endcsname
\fi
#2}}
\providecommand{\BIBdecl}{\relax}
\BIBdecl

\bibitem{ha2017sketchrnn}
D.~Ha and D.~Eck, ``A neural representation of sketch drawings,'' \emph{arXiv
  preprint arXiv:1704.03477}, 2017.

\bibitem{xu2018sketchmate}
P.~Xu, Y.~Huang, T.~Yuan, K.~Pang, Y.-Z. Song, T.~Xiang, T.~M. Hospedales,
  Z.~Ma, and J.~Guo, ``Sketchmate: Deep hashing for million-scale human sketch
  retrieval,'' in \emph{Proceedings of the IEEE Conference on Computer Vision
  and Pattern Recognition}, 2018, pp. 8090--8098.

\bibitem{liu2019sketchgan}
F.~Liu, X.~Deng, Y.-K. Lai, Y.-J. Liu, C.~Ma, and H.~Wang, ``Sketchgan: Joint
  sketch completion and recognition with generative adversarial network,'' in
  \emph{Proceedings of the IEEE Conference on Computer Vision and Pattern
  Recognition}, 2019, pp. 5830--5839.

\bibitem{sarvadevabhatla2016enabling}
R.~K. Sarvadevabhatla, J.~Kundu, and V.~Babu~R, ``Enabling my robot to play
  pictionary: Recurrent neural networks for sketch recognition,'' in
  \emph{Proceedings of the 24th ACM international conference on Multimedia},
  2016, pp. 247--251.

\bibitem{ye2016human}
Y.~Ye, Y.~Lu, and H.~Jiang, ``Human's scene sketch understanding,'' in
  \emph{Proceedings of the 2016 ACM on International Conference on Multimedia
  Retrieval}, 2016, pp. 355--358.

\bibitem{sangkloy2016sketchy}
P.~Sangkloy, N.~Burnell, C.~Ham, and J.~Hays, ``The sketchy database: learning
  to retrieve badly drawn bunnies,'' \emph{ACM Transactions on Graphics},
  vol.~35, no.~4, pp. 1--12, 2016.

\bibitem{liu2017deep}
L.~Liu, F.~Shen, Y.~Shen, X.~Liu, and L.~Shao, ``Deep sketch hashing: Fast
  free-hand sketch-based image retrieval,'' in \emph{Proceedings of the IEEE
  conference on computer vision and pattern recognition}, 2017, pp. 2862--2871.

\bibitem{shen2018zero}
Y.~Shen, L.~Liu, F.~Shen, and L.~Shao, ``Zero-shot sketch-image hashing,'' in
  \emph{Proceedings of the IEEE Conference on Computer Vision and Pattern
  Recognition}, 2018, pp. 3598--3607.

\bibitem{Collomosse_2019_CVPR}
J.~Collomosse, T.~Bui, and H.~Jin, ``Livesketch: Query perturbations for guided
  sketch-based visual search,'' in \emph{Proceedings of the IEEE Conference on
  Computer Vision and Pattern Recognition}, 2019, pp. 2879--2887.

\bibitem{Dutta_2019_CVPR}
A.~Dutta and Z.~Akata, ``Semantically tied paired cycle consistency for
  zero-shot sketch-based image retrieval,'' in \emph{Proceedings of the IEEE
  Conference on Computer Vision and Pattern Recognition}, 2019, pp. 5089--5098.

\bibitem{Dey_2019_CVPR}
S.~Dey, P.~Riba, A.~Dutta, J.~Llados, and Y.-Z. Song, ``Doodle to search:
  Practical zero-shot sketch-based image retrieval,'' in \emph{Proceedings of
  the IEEE Conference on Computer Vision and Pattern Recognition}, 2019, pp.
  2179--2188.

\bibitem{Chen_2018_CVPR}
W.~Chen and J.~Hays, ``Sketchygan: Towards diverse and realistic sketch to
  image synthesis,'' in \emph{Proceedings of the IEEE Conference on Computer
  Vision and Pattern Recognition}, 2018, pp. 9416--9425.

\bibitem{lu2018image}
Y.~Lu, S.~Wu, Y.-W. Tai, and C.-K. Tang, ``Image generation from sketch
  constraint using contextual gan,'' in \emph{Proceedings of the European
  Conference on Computer Vision}, 2018, pp. 205--220.

\bibitem{9119480}
P.~{Xu}, Z.~{Song}, Q.~{Yin}, Y.~{Song}, and L.~{Wang}, ``Deep self-supervised
  representation learning for free-hand sketch,'' \emph{IEEE Transactions on
  Circuits and Systems for Video Technology}, 2020.

\bibitem{lin2020sketch}
H.~Lin, Y.~Fu, X.~Xue, and Y.-G. Jiang, ``Sketch-bert: Learning sketch
  bidirectional encoder representation from transformers by self-supervised
  learning of sketch gestalt,'' in \emph{Proceedings of the IEEE Conference on
  Computer Vision and Pattern Recognition}, 2020, pp. 6758--6767.

\bibitem{yu2015sketch}
Q.~Yu, Y.~Yang, Y.-Z. Song, T.~Xiang, and T.~Hospedales, ``Sketch-a-net that
  beats humans,'' \emph{arXiv preprint arXiv:1501.07873}, 2015.

\bibitem{vaswani2017attention}
A.~Vaswani, N.~Shazeer, N.~Parmar, J.~Uszkoreit, L.~Jones, A.~N. Gomez,
  {\L}.~Kaiser, and I.~Polosukhin, ``Attention is all you need,''
  \emph{Advances in neural information processing systems}, vol.~30, pp.
  5998--6008, 2017.

\bibitem{battaglia2018relational}
P.~W. Battaglia, J.~B. Hamrick, V.~Bapst, A.~Sanchez-Gonzalez, V.~Zambaldi,
  M.~Malinowski, A.~Tacchetti, D.~Raposo, A.~Santoro, R.~Faulkner
  \emph{et~al.}, ``Relational inductive biases, deep learning, and graph
  networks,'' \emph{arXiv preprint arXiv:1806.01261}, 2018.

\bibitem{li2019toward}
K.~Li, K.~Pang, Y.-Z. Song, T.~Xiang, T.~M. Hospedales, and H.~Zhang, ``Toward
  deep universal sketch perceptual grouper,'' \emph{IEEE Transactions on Image
  Processing}, vol.~28, no.~7, pp. 3219--3231, 2019.

\bibitem{xu2020deep}
P.~Xu, ``Deep learning for free-hand sketch: A survey,'' \emph{arXiv preprint
  arXiv:2001.02600}, 2020.

\bibitem{krizhevsky2012imagenet}
A.~Krizhevsky, I.~Sutskever, and G.~E. Hinton, ``Imagenet classification with
  deep convolutional neural networks,'' \emph{Communications of the ACM},
  vol.~60, no.~6, pp. 84--90, 2017.

\bibitem{song2017deep}
J.~Song, Q.~Yu, Y.-Z. Song, T.~Xiang, and T.~M. Hospedales, ``Deep
  spatial-semantic attention for fine-grained sketch-based image retrieval,''
  in \emph{Proceedings of the IEEE International Conference on Computer
  Vision}, 2017, pp. 5551--5560.

\bibitem{jia2017sequential}
Q.~Jia, M.~Yu, X.~Fan, and H.~Li, ``Sequential dual deep learning with shape
  and texture features for sketch recognition,'' \emph{arXiv preprint
  arXiv:1708.02716}, 2017.

\bibitem{9068451}
L.~{Li}, C.~{Zou}, Y.~{Zheng}, Q.~{Su}, H.~{Fu}, and C.~L. {Tai},
  ``Sketch-r2cnn: An rnn-rasterization-cnn architecture for vector sketch
  recognition,'' \emph{IEEE Transactions on Visualization and Computer
  Graphics}, 2020.

\bibitem{wu2020comprehensive}
Z.~Wu, S.~Pan, F.~Chen, G.~Long, C.~Zhang, and S.~Y. Philip, ``A comprehensive
  survey on graph neural networks,'' \emph{IEEE Transactions on Neural Networks
  and Learning Systems}, pp. 1--21, 2020.

\bibitem{bruna2014spectral}
J.~Bruna, W.~Zaremba, A.~Szlam, and Y.~Lecun, ``Spectral networks and locally
  connected networks on graphs,'' in \emph{International Conference on Learning
  Representations}, 2014.

\bibitem{defferrard2016convolutional}
M.~Defferrard, X.~Bresson, and P.~Vandergheynst, ``Convolutional neural
  networks on graphs with fast localized spectral filtering,'' \emph{Advances
  in neural information processing systems}, vol.~29, pp. 3844--3852, 2016.

\bibitem{sukhbaatar2016gcn}
S.~Sukhbaatar, R.~Fergus \emph{et~al.}, ``Learning multiagent communication
  with backpropagation,'' \emph{Advances in neural information processing
  systems}, vol.~29, pp. 2244--2252, 2016.

\bibitem{kipf2017semi}
T.~N. Kipf and M.~Welling, ``Semi-supervised classification with graph
  convolutional networks,'' in \emph{International Conference on Learning
  Representations}, 2017.

\bibitem{hamilton2017inductive}
W.~Hamilton, Z.~Ying, and J.~Leskovec, ``Inductive representation learning on
  large graphs,'' in \emph{Advances in neural information processing systems},
  2017, pp. 1024--1034.

\bibitem{monti2017geometric}
F.~Monti, D.~Boscaini, J.~Masci, E.~Rodola, J.~Svoboda, and M.~M. Bronstein,
  ``Geometric deep learning on graphs and manifolds using mixture model cnns,''
  in \emph{Proceedings of the IEEE Conference on Computer Vision and Pattern
  Recognition}, 2017, pp. 5115--5124.

\bibitem{shanthamallu2019gramme}
U.~S. Shanthamallu, J.~J. Thiagarajan, H.~Song, and A.~Spanias, ``Gramme:
  Semisupervised learning using multilayered graph attention models,''
  \emph{IEEE Transactions on Neural Networks and Learning Systems}, vol.~31,
  no.~10, pp. 3977--3988, 2020.

\bibitem{edunov2018understanding}
S.~Edunov, M.~Ott, M.~Auli, and D.~Grangier, ``Understanding back-translation
  at scale,'' \emph{arXiv preprint arXiv:1808.09381}, 2018.

\bibitem{wang2019learning}
Q.~Wang, B.~Li, T.~Xiao, J.~Zhu, C.~Li, D.~F. Wong, and L.~S. Chao, ``Learning
  deep transformer models for machine translation,'' \emph{arXiv preprint
  arXiv:1906.01787}, 2019.

\bibitem{radford2018improving}
A.~Radford, K.~Narasimhan, T.~Salimans, and I.~Sutskever, ``Improving language
  understanding by generative pre-training,'' \emph{OpenAI Blog}, 2018.

\bibitem{dai2019transformer}
Z.~Dai, Z.~Yang, Y.~Yang, W.~W. Cohen, J.~Carbonell, Q.~V. Le, and
  R.~Salakhutdinov, ``Transformer-xl: Attentive language models beyond a
  fixed-length context,'' \emph{arXiv preprint arXiv:1901.02860}, 2019.

\bibitem{devlin2019bert}
J.~Devlin, M.-W. Chang, K.~Lee, and K.~Toutanova, ``Bert: Pre-training of deep
  bidirectional transformers for language understanding,'' \emph{arXiv preprint
  arXiv:1810.04805}, 2018.

\bibitem{yang2019xlnet}
Z.~Yang, Z.~Dai, Y.~Yang, J.~Carbonell, R.~Salakhutdinov, and Q.~V. Le,
  ``Xlnet: Generalized autoregressive pretraining for language understanding,''
  \emph{arXiv preprint arXiv:1906.08237}, 2019.

\bibitem{bahdanau2014neural}
D.~Bahdanau, K.~Cho, and Y.~Bengio, ``Neural machine translation by jointly
  learning to align and translate,'' \emph{arXiv preprint arXiv:1409.0473},
  2014.

\bibitem{velickovic2018graph}
P.~Veli{\v{c}}kovi{\'{c}}, G.~Cucurull, A.~Casanova, A.~Romero, P.~Li{\`{o}},
  and Y.~Bengio, ``{Graph Attention Networks},'' in \emph{International
  Conference on Learning Representations}, 2018.

\bibitem{ye2019bp}
Z.~Ye, Q.~Guo, Q.~Gan, X.~Qiu, and Z.~Zhang, ``Bp-transformer: Modelling
  long-range context via binary partitioning,'' \emph{arXiv preprint
  arXiv:1911.04070}, 2019.

\bibitem{srivastava2014dropout}
N.~Srivastava, G.~Hinton, A.~Krizhevsky, I.~Sutskever, and R.~Salakhutdinov,
  ``Dropout: A simple way to prevent neural networks from overfitting,''
  \emph{The journal of machine learning research}, vol.~15, no.~1, pp.
  1929--1958, 2014.

\bibitem{he2016deep}
K.~He, X.~Zhang, S.~Ren, and J.~Sun, ``Deep residual learning for image
  recognition,'' in \emph{Proceedings of the IEEE conference on computer vision
  and pattern recognition}, 2016, pp. 770--778.

\bibitem{Ioffe2015}
S.~Ioffe and C.~Szegedy, ``Batch normalization: Accelerating deep network
  training by reducing internal covariate shift,'' \emph{arXiv preprint
  arXiv:1502.03167}, 2015.

\bibitem{eitz2012humans}
M.~Eitz, J.~Hays, and M.~Alexa, ``How do humans sketch objects?'' \emph{ACM
  Transactions on graphics}, vol.~31, no.~4, pp. 1--10, 2012.

\bibitem{paszke2019pytorch}
A.~Paszke, S.~Gross, F.~Massa, A.~Lerer, J.~Bradbury, G.~Chanan, T.~Killeen,
  Z.~Lin, N.~Gimelshein, L.~Antiga \emph{et~al.}, ``Pytorch: An imperative
  style, high-performance deep learning library,'' in \emph{Advances in neural
  information processing systems}, 2019, pp. 8026--8037.

\bibitem{glorot2011deep}
X.~Glorot, A.~Bordes, and Y.~Bengio, ``Deep sparse rectifier neural networks,''
  in \emph{Proceedings of the fourteenth international conference on artificial
  intelligence and statistics}, 2011, pp. 315--323.

\bibitem{kingma2014adam}
D.~P. Kingma and J.~Ba, ``Adam: A method for stochastic optimization,''
  \emph{arXiv preprint arXiv:1412.6980}, 2014.

\bibitem{hochreiter1997long}
S.~Hochreiter and J.~Schmidhuber, ``Long short-term memory,'' \emph{Neural
  computation}, 1997.

\bibitem{cho2014properties}
K.~Cho, B.~Van~Merri{\"e}nboer, D.~Bahdanau, and Y.~Bengio, ``On the properties
  of neural machine translation: Encoder-decoder approaches,'' \emph{arXiv
  preprint arXiv:1409.1259}, 2014.

\bibitem{szegedy2016rethinking}
C.~Szegedy, V.~Vanhoucke, S.~Ioffe, J.~Shlens, and Z.~Wojna, ``Rethinking the
  inception architecture for computer vision,'' in \emph{Proceedings of the
  IEEE conference on computer vision and pattern recognition}, 2016, pp.
  2818--2826.

\bibitem{Sandler_2018_CVPR}
M.~Sandler, A.~Howard, M.~Zhu, A.~Zhmoginov, and L.-C. Chen, ``Mobilenetv2:
  Inverted residuals and linear bottlenecks,'' in \emph{Proceedings of the IEEE
  conference on computer vision and pattern recognition}, 2018, pp. 4510--4520.

\bibitem{huang2017densely}
G.~Huang, Z.~Liu, L.~Van Der~Maaten, and K.~Q. Weinberger, ``Densely connected
  convolutional networks,'' in \emph{Proceedings of the IEEE conference on
  computer vision and pattern recognition}, 2017, pp. 4700--4708.

\bibitem{bresson2018experimental}
X.~Bresson and T.~Laurent, ``An experimental study of neural networks for
  variable graphs,'' in \emph{International Conference on Learning
  Representations Workshop}, 2018.

\bibitem{simonyan2014very}
K.~Simonyan and A.~Zisserman, ``Very deep convolutional networks for
  large-scale image recognition,'' \emph{arXiv preprint arXiv:1409.1556}, 2014.

\bibitem{liu2020improving}
J.-J. Liu, Q.~Hou, M.-M. Cheng, C.~Wang, and J.~Feng, ``Improving convolutional
  networks with self-calibrated convolutions,'' in \emph{Proceedings of the
  IEEE Conference on Computer Vision and Pattern Recognition}, 2020, pp.
  10\,096--10\,105.

\bibitem{haase2020rethinking}
D.~Haase and M.~Amthor, ``Rethinking depthwise separable convolutions: How
  intra-kernel correlations lead to improved mobilenets,'' in \emph{Proceedings
  of the IEEE Conference on Computer Vision and Pattern Recognition}, 2020, pp.
  14\,600--14\,609.

\bibitem{pascanu2013difficulty}
R.~Pascanu, T.~Mikolov, and Y.~Bengio, ``On the difficulty of training
  recurrent neural networks,'' in \emph{International conference on machine
  learning}, 2013, pp. 1310--1318.

\bibitem{shoeybi2019megatron}
M.~Shoeybi, M.~Patwary, R.~Puri, P.~LeGresley, J.~Casper, and B.~Catanzaro,
  ``Megatron-lm: Training multi-billion parameter language models using gpu
  model parallelism,'' \emph{arXiv preprint arXiv:1909.08053}, 2019.

\bibitem{guo2019single}
X.~Guo, H.~Zhang, H.~Yang, L.~Xu, and Z.~Ye, ``A single attention-based
  combination of cnn and rnn for relation classification,'' \emph{IEEE Access},
  vol.~7, pp. 12\,467--12\,475, 2019.

\bibitem{wang2019extracting}
H.~Wang, M.~Tan, M.~Yu, S.~Chang, D.~Wang, K.~Xu, X.~Guo, and S.~Potdar,
  ``Extracting multiple-relations in one-pass with pre-trained transformers,''
  \emph{arXiv preprint arXiv:1902.01030}, 2019.

\bibitem{hendrickx2010semeval}
I.~Hendrickx, S.~N. Kim, Z.~Kozareva, P.~Nakov, D.~O. S{\'e}aghdha,
  S.~Pad{\'o}, M.~Pennacchiotti, L.~Romano, and S.~Szpakowicz, ``Semeval-2010
  task 8: Multi-way classification of semantic relations between pairs of
  nominals,'' \emph{arXiv preprint arXiv:1911.10422}, 2019.

\bibitem{devlin2018bert}
J.~Devlin, M.-W. Chang, K.~Lee, and K.~Toutanova, ``Bert: Pre-training of deep
  bidirectional transformers for language understanding,'' \emph{arXiv preprint
  arXiv:1810.04805}, 2018.

\end{thebibliography}
}

\end{document}